\documentclass[nohyperref]{article}

\usepackage{microtype}
\usepackage{graphicx}
\usepackage{subfigure}
\usepackage{booktabs} 

\usepackage[colorlinks,linkcolor=blue]{hyperref}

\usepackage[accepted]{icml2023}

\usepackage{amsmath}
\usepackage{amssymb}
\usepackage{mathtools}
\usepackage{amsthm}

\usepackage{xcolor,colortbl}
\usepackage{marvosym}
\usepackage{wrapfig}
\usepackage{caption}
\usepackage{subfigure}
\usepackage{multirow}
\usepackage{url}
\newcommand{\tabincell}[2]{\begin{tabular}{@{}#1@{}}#2\end{tabular}}
\usepackage{array}
\usepackage{booktabs}
\usepackage[title]{appendix}
\usepackage{verbatim}
\usepackage{algorithm}
\usepackage{algorithmic}
\usepackage{longtable}
\renewcommand{\algorithmicrequire}{\textbf{Input:}} 
\renewcommand{\algorithmicensure}{\textbf{Output:}}
\usepackage{rotating}

\usepackage{bbding}
\usepackage{threeparttable}
\usepackage{pifont}
\usepackage{arydshln}

\usepackage[capitalize,noabbrev]{cleveref}

\theoremstyle{plain}

\theoremstyle{definition}

\theoremstyle{remark}

\usepackage[textsize=tiny]{todonotes}

\icmltitlerunning{BiBench: Benchmarking and Analyzing Network Binarization}

\begin{document}

\twocolumn[
\icmltitle{BiBench: Benchmarking and Analyzing Network Binarization}



\icmlsetsymbol{equal}{*}

\begin{icmlauthorlist}
\vspace{-0.12in}
\icmlauthor{Haotong Qin}{equal,sch1,sch2}
\icmlauthor{Mingyuan Zhang}{equal,sch3}
\icmlauthor{Yifu Ding}{sch1}
\icmlauthor{Aoyu Li}{sch3}
\icmlauthor{Zhongang Cai}{sch3}\\
\icmlauthor{Ziwei Liu}{sch3}
\icmlauthor{Fisher Yu}{sch2}
\icmlauthor{Xianglong Liu\textsuperscript{\Letter}}{sch1}
\vspace{-0.12in}
\end{icmlauthorlist}

\icmlaffiliation{sch1}{Beihang University}
\icmlaffiliation{sch2}{ETH Zürich}
\icmlaffiliation{sch3}{Nanyang Technological University}

\icmlcorrespondingauthor{\textsuperscript{\Letter} Xianglong Liu}{xlliu@buaa.edu.cn}

\icmlkeywords{Network Binarization, Model Compression, Deep Learning}

\vskip 0.3in
]



\printAffiliationsAndNotice{\icmlEqualContribution} 

\begin{abstract}
Network binarization emerges as one of the most promising compression approaches offering extraordinary computation and memory savings by minimizing the bit-width. However, recent research has shown that applying existing binarization algorithms to diverse tasks, architectures, and hardware in realistic scenarios is still not straightforward. Common challenges of binarization, such as accuracy degradation and efficiency limitation, suggest that its attributes are not fully understood. 
To close this gap, we present \textbf{BiBench}, a rigorously designed benchmark with in-depth analysis for network binarization. We first carefully scrutinize the requirements of binarization in the actual production and define evaluation tracks and metrics for a comprehensive and fair investigation. Then, we evaluate and analyze a series of milestone binarization algorithms that function at the operator level and with extensive influence.
Our benchmark reveals that 1) the binarized operator has a crucial impact on the performance and deployability of binarized networks; 2) the accuracy of binarization varies significantly across different learning tasks and neural architectures; 3) binarization has demonstrated promising efficiency potential on edge devices despite the limited hardware support. The results and analysis also lead to a promising paradigm for accurate and efficient binarization. We believe that BiBench will contribute to the broader adoption of binarization and serve as a foundation for future research.
The code for our BiBench is released \href{https://github.com/htqin/BiBench}{here}.
\vspace{-0.1in}

\end{abstract}

\section{Introduction}


\begin{figure*}[t]
\centering
\includegraphics[width=\linewidth]{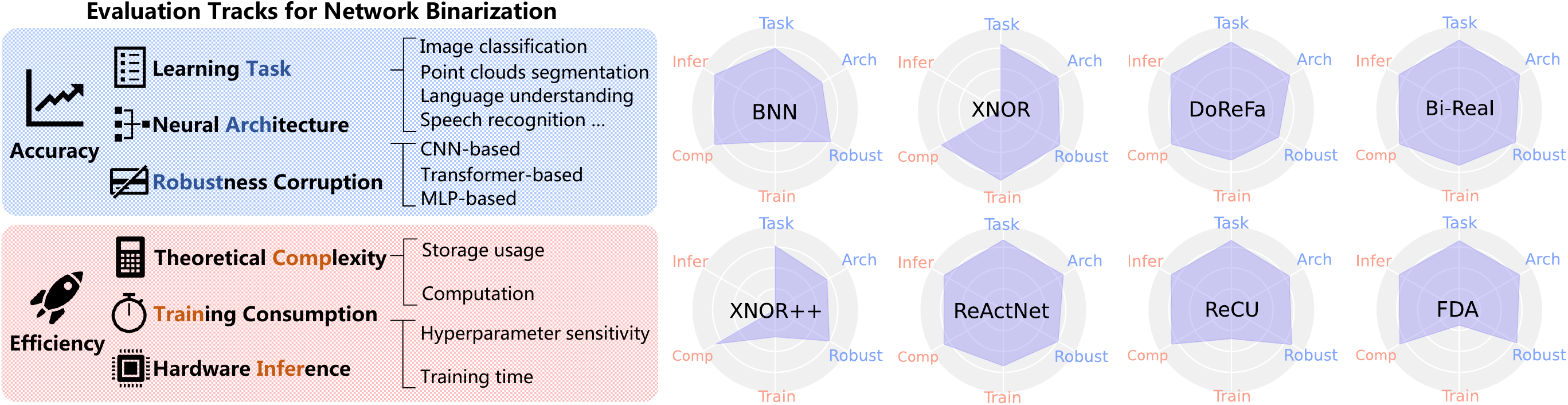}
\vspace{-0.2in}
\caption{Evaluation tracks of BiBench.
Our benchmark evaluates the performance of binarization algorithms on a range of comprehensive evaluation tracks, including: ``Learning Task'', ``Neural Architecture'', ``Corruption Robustness'', ``Training Consumption'', ``Theoretical Complexity'', and ``Hardware Inference''.}
\label{fig:ovewview}
\end{figure*}

The rising of deep learning leads to the persistent contradiction between larger models and the limitations of deployment resources. Compression technologies have been widely studied to address this issue, including quantization~\citep{gong2014compressing,wu2016quantized,Vanhoucke2011ImprovingTS,gupta2015deep}, pruning~\citep{han2015learning, han2016deep, he2017channel}, distillation~\citep{hinton2015distilling,xu2018training,chen2018darkrank,Yim2017A,zagoruyko2017paying}, lightweight architecture design~\citep{mobilenet,mobilenet_v2,shufflenet,shufflenet_v2}, and low-rank decomposition~\citep{denton2014exploiting,lebedev2015speeding,jaderberg2014speeding,lebedev2016fast}. These technologies are essential for the practical application of deep learning.

As a compression approach that reduces the bit-width to 1-bit, network binarization is regarded as the most aggressive quantization technology~\citep{rusci2020memory,choukroun2019low,ijcai2022p603,shang2022network,zhang2022pokebnn,bethge2020meliusnet,bethge2019back,martinez2019training,helwegen2019latent}. The binarized models take little storage and memory and accelerate the inference by efficient bitwise operations. Compared to other compression technologies like pruning and architecture design, network binarization has potent topological generics, as it only applies to parameters. As a result, it is widely studied in academic research as a standalone compression technique rather than just a 1-bit specialization of quantization~\citep{Gong:iccv19,gholamisurvey}. Some state-of-the-art (SOTA) binarization algorithms have even achieved full-precision performance with binarized models on large-scale tasks~\citep{deng2009imagenet,liu2020reactnet}.

However, existing network binarization is still far from practical, and two worrisome trends appear in current research:

\textbf{\textit{Trend-1}: Accuracy comparison scope is limited.} 
In recent research, several image classification tasks (such as CIFAR-10 and ImageNet) have become standard options for comparing accuracy among different binarization algorithms.
While this helps to clearly and fairly compare accuracy performance, it causes most binarization algorithms to be only engineered for image inputs (2D visual modality), and their insights and conclusions are rarely verified in other modalities and tasks. The use of monotonic tasks also hinders a comprehensive evaluation from an architectural perspective. 
Furthermore, data noise, such as corruption~\citep{hendrycks2018benchmarking}, is a common problem on low-cost edge devices and is widely studied in compression, but this situation is hardly considered existing binarization algorithms.

\textbf{\textit{Trend-2}: Efficiency analysis remains theoretical.} Network binarization is widely recognized for its significant storage and computation savings, with theoretical savings of up to 32$\times$ and 64$\times$ for convolutions, respectively~\citep{rastegari2016xnor,bai2021binarybert}. However, these efficiency claims lack experimental evidence due to the lack of hardware library support for deploying binarized models on real-world hardware. Additionally, the training efficiency of binarization algorithms is often ignored in current research, leading to negative phenomena during the training of binary networks, such as increased demand for computation resources and time consumption, sensitivity to hyperparameters, and the need for detailed optimization tuning.

This paper presents \textbf{BiBench}, a network \textbf{bi}narization \textbf{bench}mark designed to evaluate binarization algorithms comprehensively in terms of accuracy and efficiency (Table~\ref{tab:overview}). Using BiBench, we select 8 representative binarization algorithms that are extensively influential and function at the operator level (details and selection criteria are in Appendix~\ref{app:algo}) and benchmark algorithms on 9 deep learning datasets, 13 neural architectures, 2 deployment libraries, 14 hardware chips, and various hyperparameter settings. We invested approximately 4 GPU years of computation time in creating BiBench, intending to promote a comprehensive evaluation of network binarization from both accuracy and efficiency perspectives. We also provide in-depth analysis of the benchmark results, uncovering insights and offering suggestions for designing practical binarization algorithms.

We emphasize that our BiBench includes the following significant contributions: 
(1) \textit{The \textbf{first} systematic benchmark enables a new view to quantitatively evaluate binarization algorithms at the operator level.} BiBench is the first effort to facilitate systematic and comprehensive comparisons between binarized algorithms. It provides a brand new perspective to decouple the binarized operators from the neural architectures for quantitative evaluations at the generic operator level. 
(2) \textit{Revealing a practical binarized operator design paradigm.} BiBench reveals a practical paradigm of binarized operator designing. Based on the systemic and quantitative evaluation, superior techniques for more satisfactory binarization operators can emerge, which is essential for pushing binarization algorithms to be accurate and efficient. 
A more detailed discussion is in Appendix~\ref{subsec:DiscussionofNovelty}.

\section{Background}
\label{sec:background}

\begin{table*}[t]
\begin{center}
\caption{
Comparison between BiBench and existing binarization works along evaluation tracks.}
\label{tab:overview}
\vspace{-0.15in}
\setlength{\tabcolsep}{3.2mm}
{\begin{threeparttable}
\begin{tabular}{lccccccccc}
\toprule
\multirow{2.4}{*}{Algorithm} &  \multicolumn{3}{c}{Technique} & \multicolumn{3}{c}{Accurate Binarization} & \multicolumn{3}{c}{Efficient Binarization} \\
\cmidrule(r){2-4}\cmidrule(r){5-7}\cmidrule(r){8-10}
& $s$ & $\tau$ & $g$ & {\#Task} & {\#Arch} & Robust & Train & Comp & Infer\\
\midrule
BNN~\citep{courbariaux2016binarized} & $\times$ & $\times$ & $\surd$ & 3 & 3 & * &  $\surd$ & $\surd$ & $\surd$ \\
XNOR~\citep{rastegari2016xnor} & $\surd$ & $\times$ & $\times$ & 2 & 3 &  * &  $\surd$ & $\surd$ &  $\surd$ \\
DoReFa~\citep{Dorefa-Net} & $\surd$ & $\times$ & $\times$ & 2 & 2 & * &  $\times$ & $\surd$ & $\times$  \\
Bi-Real~\citep{liu2018bi} & $\times$ & $\times$ & $\surd$ & 1 & 2 &  $\times$ &  $\times$ & $\surd$ &  $\times$ \\
XNOR++~\citep{bulat2019xnor} & $\surd$ & $\times$ & $\times$ & 1 & 2 &  $\times$ &  $\times$ &  $\times$ & $\times$  \\
ReActNet~\citep{liu2020reactnet} & $\times$ & $\surd$ & $\times$ & 1 & 2 &  $\times$ &  $\times$ & $\surd$ &  $\times$ \\
ReCU~\citep{xu2021recu} & $\times$ & $\surd$ & $\surd$ & 2 & 4 &  $\times$ &  $\times$ &  $\times$ & $\times$  \\
FDA~\citep{xu2021learning} & $\times$ & $\times$ & $\surd$ & 1 & 6 &  $\times$ &  $\times$ &  $\times$ & $\times$  \\
\midrule
\textit{Our Benchmark (\textbf{BiBench})} & $\surd$ & $\surd$ & $\surd$ & \textbf{9} & \textbf{13} & \textbf{$\surd$} & \textbf{$\surd$} & \textbf{$\surd$} & \textbf{$\surd$} \\%
\bottomrule
\end{tabular}
\begin{tablenotes}
\item[1] ``$\surd$" and ``$\times$" indicates the track is considered in the original binarization algorithm, while ``*" indicates only being studied in other related studies. ``$s$", ``$\tau$", or ``$g$" indicates ``scaling factor", ``parameter redistribution", or ``gradient approximation" techniques proposed in this work, respectively. And we also present a more detailed list of these techniques (Table~\ref{tab:algos_detail}) for binarization algorithms in Appendix~\ref{app:algo}.
\end{tablenotes}
\end{threeparttable}}
\end{center}
\end{table*}

\subsection{Network Binarization}
\label{subsec:NetworkBinarization}

Binarization compresses weights $\boldsymbol{w}\in \mathbb{R}^{c_\text{in}\times c_\text{out}\times k\times k}$ and activations $\boldsymbol{a}\in \mathbb{R}^{c_\text{in}\times w\times h}$ to 1-bit in computationally dense convolution, where $c_\text{in}$, $k$, $c_\text{out}$, $w$, and $h$ denote the input channel, kernel size, output channel, input width, and input height. The computation can be expressed as
\begin{equation}
\label{eq:quantized_net}
\boldsymbol o = \alpha\operatorname{popcount}\left(\operatorname{xnor}\left(\operatorname{sign}(\boldsymbol{a}), \operatorname{sign}(\boldsymbol{w})\right)\right),
\end{equation}
where $\boldsymbol{o}$ denotes the outputs and $\alpha\in \mathbb{R}^{c_\text{out}}$ denotes the optional scaling factor calculated as $\alpha=\frac{\|w\|}{n}$~\citep{courbariaux2016binarized,rastegari2016xnor}, $\operatorname{xnor}$ and $\operatorname{popcount}$ are bitwise instructions defined as~\citep{arm64,x86}. The $\operatorname{popcount}$ counts the number of bits with the "one" value in the input vector and writes the result to the targeted register.
While binarized parameters of networks offer significant compression and acceleration benefits, their limited representation can lead to decreased accuracy. Various algorithms have been proposed to address this issue to improve the accuracy of binarized networks~\citep{yuan2021comprehensive}.

The majority of binarization algorithms aim to improve binarized operators (as Eq.~(\ref{eq:quantized_net}) shows), which play a crucial role in the optimization and hardware efficiency of binarized models~\citep{alizadeh2018a,geiger2020larq}. These operator improvements are also flexible across different neural architectures and learning tasks, demonstrating the generalizability of bit-width compression~\citep{wang2020bidet,ijcai2022p603,zhao2022pb}. Our BiBench considers 8 extensively influential binarization algorithms that focus on improving operators and can be broadly classified into three categories: scaling factors, parameter redistribution, and gradient approximation~\citep{courbariaux2016binarized,rastegari2016xnor,Dorefa-Net,liu2018bi,bulat2019xnor,liu2020reactnet,xu2021recu,xu2021learning}.
Note that for selected binarization algorithms, the techniques requiring specified local structures or training pipelines are excluded for fairness, \textit{i.e.}, the bi-real shortcut of Bi-Real~\citep{Bi-Real} and duplicate activation of ReActNet~\citep{liu2020reactnet} in CNN neural architectures.
See Appendix~\ref{app:algo} for more information on the algorithms in our BiBench.

\subsection{Challenges for Binarization}

Since around 2015, network binarization has garnered significant attention in various fields of research, including but not limited to vision and language understanding. However, several challenges still arise during the production and deployment of binarized networks in practice. 
The goal of binarization production is to train accurate binarized networks that are resource-efficient. Some recent studies have demonstrated that the performance of binarization algorithms on image classification tasks may not always generalize to other learning tasks and neural architectures~\citep{qin2020bipointnet,wang2020bidet,qin2021bibert,liu2022bit}. In order to achieve higher accuracy, some binarization algorithms may require several times more training resources compared to full-precision networks.
Ideally, binarized networks should be hardware-friendly and robust when deployed on edge devices. However, most mainstream inference libraries do not currently support the deployment of binarized networks on hardware~\citep{tensorrt,acl,snpe}, which limits the performance of existing binarization algorithms in practice. In addition, the data collected by low-cost devices in natural edge scenarios is often of low quality and even be corrupted, which can negatively impact the robustness of binarized models~\citep{lin2018defensive,ye2019adversarial,cygert2021robustness}. However, most existing binarization algorithms do not consider corruption robustness when designed.

\section{BiBench: Tracks and Metrics}



In this section, we present BiBench, a benchmark for accurate and efficient network binarization. Our evaluation consists of 6 tracks and corresponding metrics, as shown in Figure~\ref{fig:ovewview}, which address the practical challenges of producing and deploying binarized networks. Higher scores on these metrics indicate better performance.

\subsection{Towards Accurate Binarization}
\label{subsec:TowardsAccurateBinarization}

In our BiBench, the evaluation tracks for accurate binarization are ``Learning Task", ``Neural Architecture" (for production), and ``Corruption Robustness" (for deployment).

\ding{172}\ \textbf{Learning Task}.
We comprehensively evaluate network binarization algorithms using 9 learning tasks across 4 different data modalities. For the widely-evaluated 2D visual modality tasks, we include image classification on CIFAR-10~\citep{CIFAR} and ImageNet~\citep{krizhevsky2012imagenet} datasets, as well as object detection on PASCAL VOC~\citep{hoiem2009pascal} and COCO~\citep{lin2014microsoft} datasets. In the 3D visual modality tasks, we evaluate the algorithm on ModelNet40 classification~\citep{wu20153d} and ShapeNet segmentation~\citep{chang2015shapenet} datasets of 3D point clouds. For the textual modality tasks, we use the natural language understanding in the GLUE benchmark~\citep{wang2018glue}. For the speech modality tasks, we evaluate the algorithms on the Speech Commands KWS dataset~\citep{warden2018speech}. See Appendix.~\ref{app:task} for more details on the tasks and datasets.

To evaluate the performance of a binarization algorithm on this track, we use the accuracy of full-precision models as a baseline and calculate the mean relative accuracy for all architectures on each task. The Overall Metric (OM) for this track is then calculated as the quadratic mean of the relative accuracies across all tasks~\citep{curtis2000quadratic}. The equation for this evaluation metric is as follows:
\begin{equation}
\label{eq:track1}
\textrm{OM}_\textrm{task}=\sqrt{\frac{1}{N}\sum\limits_{i=1}^N\mathbb{E}^2\left(\frac{\boldsymbol{A}_{\textrm{task}_i}^{bi}}{\boldsymbol{A}_{\textrm{task}_i}}\right)},
\end{equation}
where $\boldsymbol{A}_{\textrm{task}_i}^{bi}$ and $\boldsymbol{A}_{\textrm{task}_i}$ denote the accuracy of the binarized and full-precision models on the $i$-th task, respectively, $N$ is the number of tasks, and $\mathbb{E}(\cdot)$ is the mean operation. 

Note that the quadratic mean form is used uniformly in BiBench to unify all overall metrics of tracks, which helps prevent certain poor performers from disproportionately influencing the metric and allows for a more accurate measure of overall performance on each track.

\ding{173}\ \textbf{Neural Architecture}.
We evaluate various neural architectures, including mainstream CNN-based, transformer-based, and MLP-based architectures, to assess the generalizability of binarization algorithms from the perspective of neural architecture. Specifically, we use standard ResNet-18/20/34~\citep{he2016deep} and VGG~\citep{simonyan2015very} to evaluate CNN architectures, and apply the Faster-RCNN~\citep{NIPS2015_5638} and SSD300~\citep{liu2016ssd} frameworks as detectors. To evaluate transformer-based architectures, we binarize BERT-Tiny4/Tiny6/Base~\citep{kenton2019bert} with the bi-attention mechanism~\citep{qin2021bibert} for convergence. We also evaluate MLP-based architectures, including PointNet$_\text{vanilla}$ and PointNet~\citep{qi2017pointnet} with EMA aggregator~\citep{qin2020bipointnet}, FSMN~\citep{zhang2015feedforward}, and Deep-FSMN~\citep{zhang2018deep}, due to their linear unit composition. Detailed descriptions of these architectures can be found in Appendix~\ref{app:arch}.

Similar to the overall metric for learning task track, we build the overall metric for neural architecture track:
\begin{equation}
\scriptsize
\textrm{OM}_\textrm{arch}=\sqrt{\frac{1}{3}\left(\mathbb{E}^2\left(\frac{\boldsymbol{A}_{\textrm{CNN}}^{bi}}{\boldsymbol{A}_{\textrm{CNN}}}\right)+\mathbb{E}^2\left(\frac{\boldsymbol{A}_{\textrm{Transformer}}^{bi}}{\boldsymbol{A}_{\textrm{Transformer}}}\right)+\mathbb{E}^2\left(\frac{\boldsymbol{A}_{\textrm{MLP}}^{bi}}{\boldsymbol{A}_{\textrm{MLP}}}\right)\right) }.
\end{equation}

\ding{174}\ \textbf{Corruption Robustness}.
The corruption robustness of binarization on deployment is important to handle scenarios such as perceptual device damage, a common issue with low-cost equipment in real-world implementations. To assess the robustness of binarized models to corruption of 2D visual data, we evaluate algorithms on the CIFAR10-C~\citep{hendrycks2018benchmarking} benchmark.

Therefore, we evaluate the performance of binarization algorithms on corrupted data compared to normal data using the corruption generalization gap~\citep{zhang2022delving}:
\begin{equation}
\boldsymbol{G}_{\textrm{task}_i} = \boldsymbol{A}_{\textrm{task}_i}^{\textrm{norm}} - \boldsymbol{A}_{\textrm{task}_i}^{\textrm{corr}},
\end{equation}
where $\boldsymbol{A}^{\textrm{corr}}_{\textrm{task}i}$ and $\boldsymbol{A}^{\textrm{norm}}{\textrm{task}_i}$ denote the accuracy results under all architectures on the $i$-th corruption task and corresponding normal task, respectively. The overall metric on this track is then calculated by
\begin{equation}
\textrm{OM}_\textrm{robust}=\sqrt{\frac{1}{C}\sum\limits_{i=1}^C\mathbb{E}^2\left(\frac{\boldsymbol{G}_{\textrm{task}_i}}{\boldsymbol{G}_{\textrm{task}_i}^{bi}}\right) }.
\end{equation}

\subsection{Towards Efficient Binarization}

We evaluate the efficiency of network binarization in terms of ``Training Consumption" for production, ``Theoretical Complexity" and ``Hardware Inference" for deployment.

\ding{175}\ \textbf{Training Consumption}.
We consider the occupied training resource and the hyperparameter sensitivity of binarization algorithms, which impact the consumption of a single training and the overall tuning process. To evaluate the ease of tuning binarization algorithms to optimal performance, we train their binarized networks with various hyperparameter settings, including different learning rates, learning rate schedulers, optimizers, and even random seeds. We align the epochs for binarized and full-precision networks and compare their consumption and time.

The metric used to evaluate the training consumption track is based on both training time and sensitivity to hyperparameters. For one binarization algorithm, we have
\begin{equation}
\textrm{OM}_\textrm{train}=\sqrt{\frac{1}{2}\left(\mathbb{E}^2\left(\frac{\boldsymbol{T}_\textrm{train}}{\boldsymbol{T}_\textrm{train}^{bi}}\right)
+\mathbb{E}^2\left(\frac{\operatorname{std}(\boldsymbol{A}_\textrm{hyper})}{\operatorname{std}(\boldsymbol{A}^{bi}_\textrm{hyper})}\right)\right)},
\end{equation}
where the set $\boldsymbol{T}_\textrm{train}$ represents the time spent on a single training instance, $\boldsymbol{A}_\textrm{hyper}$ is the set of results obtained using different hyperparameter configurations, and $\operatorname{std}(\cdot)$ calculates standard deviation values.

\ding{176}\ \textbf{Theoretical Complexity}.
To evaluate complexity, we compute the compression and speedup ratios before and after binarization on architectures such as ResNet18.

The evaluation metric is based on model size (MB) and computational floating-point operations (FLOPs) savings during inference. Binarized parameters occupy 1/32 the storage of their 32-bit floating-point counterparts~\citep{rastegari2016xnor}. Binarized operations, where a 1-bit weight is multiplied by a 1-bit activation, take approximately 1*1/64 FLOPs on a CPU with 64-bit instruction size~\citep{Dorefa-Net,liu2018bi,li2019additive}. The compression ratio $\boldsymbol{r}_{c}$ and speedup ratio $\boldsymbol{r}_{s}$ are
\begin{equation}
\begin{aligned}
\boldsymbol{r}_{c}=&\frac{|\boldsymbol{M}|_{\ell0}}{\frac{1}{32}\left(|\boldsymbol{M}|_{\ell0}-|\boldsymbol{\hat{M}}|_{\ell0}\right)+|\boldsymbol{\hat{M}}|_{\ell0}},\\
\boldsymbol{r}_{s}=&\frac{\operatorname{FLOPs}_{\boldsymbol{M}}}{\frac{1}{64}\left(\operatorname{FLOPs}_{\boldsymbol{M}}-\operatorname{FLOPs}_{\boldsymbol{\hat{M}}}\right)+\operatorname{FLOPs}_{\boldsymbol{\hat{M}}}},
\end{aligned}
\end{equation}
where $\boldsymbol{M}$ and $\boldsymbol{\hat{M}}$ are the number of full-precision parameters that remains in the original and binarized models, respectively, and $\operatorname{FLOPs}_{\boldsymbol{M}}$ and $\operatorname{FLOPs}_{\boldsymbol{\hat{M}}}$ represent the computation related to these parameters. The overall metric for theoretical complexity is
\begin{equation}
\textrm{OM}_\textrm{comp}=\sqrt{\frac{1}{2}\left(\mathbb{E}^2(\boldsymbol{r}_{c})+\mathbb{E}^2(\boldsymbol{r}_{s})\right)}.
\end{equation}

\begin{table*}[!t]
\centering
\caption{Accuracy benchmark for network binarization. Blue: best in a row. Red: worst in a row.}
\label{tab:acc1}
\vspace{-0.15in}
\setlength{\tabcolsep}{2.15mm}
{
\begin{tabular}{llcccccccc}
\toprule
\multirow{2}{*}{Track} & \multirow{2}{*}{Metric} & \multicolumn{8}{c}{Binarization Algorithm} \\
 & & {BNN} & {XNOR} & {DoReFa} & {Bi-Real} & {XNOR++} & {ReActNet} & {ReCU} & {FDA} \\
\midrule
\multirow{9}{*}{\tabincell{l}{Learning\\Task (\%)}} & CIFAR10 & 94.54 & 94.73 & 95.03 & 95.61 & \cellcolor{red!10}94.52 & 95.92 & \cellcolor{blue!10}96.72 & 94.66 \\
 & ImageNet & 75.81 & 77.24 & 76.61 & 78.38 & \cellcolor{red!10}75.01 & \cellcolor{blue!10}78.64 & 77.98 & 78.15 \\
 & VOC07 & 76.97 & 74.61 & 76.35 & 80.07 & \cellcolor{red!10} 74.41 & 81.38 & \cellcolor{blue!10} 81.65 & 79.02\\
 & COCO17 & 77.94 &	\cellcolor{red!10} 75.37 &	78.31 &	81.62 &	79.41 &	83.82 &	\cellcolor{blue!10} 85.66 & 82.35 \\
 & ModelNet40 & \cellcolor{red!10}54.19	& 93.86	& 93.74	& 93.23 &	85.20 &	92.41 &	\cellcolor{blue!10}95.07 &	94.38 \\
 & ShapeNet & 48.96	&\cellcolor{blue!10}73.62&	70.79&	68.13&	41.16&	68.51&	\cellcolor{red!10}40.65&	71.16\\
 & GLUE &  49.75	& 59.63	& 66.60	& 69.42	& \cellcolor{red!10}49.33	& 67.64 &	50.66	& \cellcolor{blue!10}70.61 \\
 & SpeechCom. & 75.03 & 76.93 & 76.64 & \cellcolor{blue!10} 82.42 & \cellcolor{red!10} 68.65 & 81.86 & 76.98 & 77.90 \\
 & \textbf{OM$_\textrm{task}$} & \cellcolor{red!25}70.82	&78.97	&79.82	&81.63	&72.89	& \cellcolor{blue!25}81.81&	77.96	&81.49 \\
\midrule
\multirow{4}{*}{\tabincell{l}{Neural\\Architecture (\%)}} & CNNs & \cellcolor{red!10}72.90	& 83.74 &	83.86 &	85.02 &	78.95 &	86.20 &	83.50 &	\cellcolor{blue!10}86.34\\
 & Transformers & 49.75	& 59.63	& 66.60	& 69.42	& \cellcolor{red!10}49.33	& 67.64 &	50.66	& \cellcolor{blue!10}70.61 \\
 & MLPs & \cellcolor{red!10}64.61 &	85.40 &	85.19 &	\cellcolor{blue!10}87.83 &	76.92 &	87.13 &	86.02 &	86.14\\
 & \textbf{OM$_\textrm{arch}$} & \cellcolor{red!25}63.15	& 77.16 &	79.01 &	81.16 &	69.72 &	80.82 &	75.14 &	\cellcolor{blue!25}81.36 \\
\midrule
\multirow{2}{*}{\tabincell{l}{Robustness\\Corruption (\%)}} & CIFAR10-C& 95.26	&100.97	&\cellcolor{red!10}81.43 &	96.56&	92.69&	94.01&	\cellcolor{blue!10}103.29 & 98.35 \\
 & \textbf{OM$_\textrm{corr}$} & 95.26	&100.97	&\cellcolor{red!25}81.43 &	96.56&	92.69&	94.01&	\cellcolor{blue!25}103.29 & 98.35\\
\bottomrule
\end{tabular}}
\end{table*}

\ding{177}\ \textbf{Hardware Inference}.
As binarization is not widely supported in hardware deployment, only two inference libraries, Larq's Compute Engine~\citep{geiger2020larq} and JD's daBNN~\citep{zhang2019dabnn}, can deploy and evaluate binarized models on ARM hardware in practice. We focus on ARM CPU inference on mainstream hardware for edge scenarios, such as HUAWEI Kirin, Qualcomm Snapdragon, Apple M1, MediaTek Dimensity, and Raspberry Pi (details in Appendix~\ref{app:hardware}).

For a given binarization algorithm, we use the savings in storage and inference time under different inference libraries and hardware as evaluation metrics:
\begin{equation}
\textrm{OM}_\textrm{infer}=\sqrt{\frac{1}{2}\left(\mathbb{E}^2\left(\frac{\boldsymbol{T}_\textrm{infer}}{\boldsymbol{T}_\textrm{infer}^{bi}}\right) +\mathbb{E}^2\left(\frac{\boldsymbol{S}_\textrm{infer}}{\boldsymbol{S}_\textrm{infer}^{bi}}\right)\right)},
\end{equation}
where $\boldsymbol{T}_\textrm{infer}$ is the inference time and $\boldsymbol{S}_\textrm{infer}$ is the storage used on different devices.

\section{BiBench Implementation}
\label{sec:bibench_eval}

This section presents the implementation details, training, and inference pipelines of BiBench.

\noindent\textbf{Implementation details.} 
BiBench is implemented using the PyTorch~\citep{paszke2019pytorch} package. The definitions of the binarized operators are contained in individual, separate files, enabling the flexible replacement of the corresponding operator in the original model when evaluating different tasks and architectures. When deployed, well-trained binarized models for a particular binarization algorithm are exported to the Open Neural Network Exchange (ONNX) format~\citep{onnxruntime} and provided as input to the appropriate inference libraries (if applicable for the algorithm).

\noindent\textbf{Training and inference pipelines.}
\textit{Hyperparameters}: Binarized networks are trained for the same number of epochs as their full-precision counterparts. Inspired by the results in Section~\ref{sec:consumption}, we use the Adam optimizer for all binarized models for well converging. The default initial learning rate is $1e-3$ (or $0.1\times$ the default learning rate), and the learning rate scheduler is CosineAnnealingLR~\citep{loshchilov2017sgdr}.
\textit{Architecture}: BiBench follows the original architectures of full-precision models, binarizing their convolution, linear, and multiplication units with the selected binarization algorithms. Hardtanh is uniformly used as the activation function to prevent all-one features.
\textit{Pretraining}: All binarization algorithms use finetuning. For each one, all binarized models are initialized using the same pre-trained model for the specific neural architecture and learning task to eliminate inconsistency at initialization.

\section{BiBench Evaluation and Analysis}
This section presents and analyzes the evaluation results in BiBench. The main accuracy results are in Table~\ref{tab:acc1}, and the efficiency results are in Table~\ref{tab:eff}. Details are in Appendix~\ref{sec:FullResults}.

\subsection{Accuracy Tracks}
The accuracy results for network binarization are presented in Table~\ref{tab:acc1}. These results were obtained using the metrics defined in Section~\ref{subsec:TowardsAccurateBinarization} for each accuracy-related track.

\subsubsection{Learning Task: Performance Varies Greatly by Algorithms and Modalities}
We present the evaluation results of binarization on various learning tasks. In addition to the overall metric $\textrm{OM}_\textrm{task}$, we also provide the relative accuracy of binarized networks compared to their full-precision counterparts.

\textbf{The impact of binarized operators is crucial and significant}. 
With fully unified training pipelines and architectures, a substantial variation in performance appears among binarization algorithms across every learning task. For example, the SOTA FDA algorithm exhibits a 21.3\% improvement in accuracy on GLUE datasets compared to XNOR++, and the difference is even greater at 33.0\% on ShapeNet between XNOR and ReCU. This suggests that binarized operators play a crucial role in the learning task track, and their importance is also confirmed in other tracks.

\textbf{Binarization algorithms vary greatly under different data modalities}. 
When comparing various learning tasks, it is notable that binarized networks suffer a significant drop in accuracy on the language understanding GLUE benchmark but can approach full-precision performance on the ModelNet40 point cloud classification task. This and similar phenomena suggest that the direct transfer of binarization insights across learning tasks is non-trivial.

For overall performance, both ReCU and ReActNet have high accuracy across various learning tasks. While ReCU performs best on most individual tasks, ReActNet ultimately stands out in the overall metric comparison. Both algorithms apply reparameterization in the forward propagation and gradient approximation in the backward propagation.

\subsubsection{Neural Architecture: Binarization on Transformers Is Challenging}
\label{subsubsec:NeuralArchitectureBinarizationonTransformers}
\textbf{Binarization exhibits a clear advantage on CNN- and MLP-based architectures compared to transformer-based ones}. Advanced binarization algorithms can achieve 78\%-86\% of full-precision accuracy on CNNs, and binarized networks with MLP architectures can even approach full-precision performance (\textit{e.g.}, Bi-Real 87.83\%). In contrast, transformer-based architectures suffer significant performance degradation when binarized, and none of the algorithms achieve an overall accuracy higher than 70\%.

\begin{table*}[t]
\centering
\caption{Efficiency benchmark for network binarization. Blue: best in a row. Red: worst in a row.}
\label{tab:eff}
\vspace{-0.15in}
\setlength{\tabcolsep}{2.4mm}
{
\begin{tabular}{llcccccccc}
\toprule
\multirow{2}{*}{Track} & \multirow{2}{*}{Metric} & \multicolumn{8}{c}{Binarization Algorithm} \\
 & & {BNN} & {XNOR} & {DoReFa} & {Bi-Real} & {XNOR++} & {ReActNet} & {ReCU} & {FDA} \\
\midrule
\multirow{3}{*}{\tabincell{l}{Training\\Consumption\\(\%)}} & Sensitivity & \cellcolor{red!10}27.28 & \cellcolor{blue!10}175.53 & 113.40 & 144.59 & 28.66 & 146.06 & 53.33 & 36.62 \\
& Time Cost & 82.19 & 71.43 &	\cellcolor{blue!10}76.92&	45.80&	68.18	&	45.45& 58.25	&\cellcolor{red!10}20.62 \\
 & \textbf{OM$_\textrm{train}$} & 61.23&	\cellcolor{blue!25}134.00&	96.89&	107.25&	52.30&	108.16&	55.84&	\cellcolor{red!25}29.72 \\
\midrule
\multirow{3}{*}{\tabincell{l}{Theoretical\\Complexity\\($\times$)}} & Speedup & \cellcolor{blue!10}12.60 & \cellcolor{red!10}12.26 & 12.37 & 12.37 & \cellcolor{red!10}12.26 & \cellcolor{red!10}12.26 & 12.37 & 12.37  \\
 & Compression & \cellcolor{blue!10}13.27 & 13.20 & 13.20 & 13.20 & \cellcolor{red!10}13.16 & 13.20 & 13.20 & 13.20 \\
 & \textbf{OM$_\textrm{comp}$} & \cellcolor{blue!25}12.94 & 12.74 & 12.79 & 12.79 & \cellcolor{red!25}12.71 & 12.74 & 12.79 & 12.79 \\
\midrule
\multirow{3}{*}{\tabincell{l}{Hardware\\Deployment\\($\times$)}} & Speedup & \cellcolor{blue!10}5.45 & \cellcolor{red!10}False & \cellcolor{blue!10}5.45 & \cellcolor{blue!10}5.45 & \cellcolor{red!10}False & 4.89 & \cellcolor{blue!10}5.45 & \cellcolor{blue!10}5.45 \\
 & Compression & \cellcolor{blue!10}15.62 & \cellcolor{red!10}False & \cellcolor{blue!10}15.62 & \cellcolor{blue!10}15.62 & \cellcolor{red!10}False & 15.52 & \cellcolor{blue!10}15.62 & \cellcolor{blue!10}15.62 \\
 & \textbf{OM$_\textrm{infer}$} & \cellcolor{blue!25}11.70 & \cellcolor{red!25}False & \cellcolor{blue!25}11.70 & \cellcolor{blue!25}11.70 & \cellcolor{red!25}False & 11.51 & \cellcolor{blue!25}11.70 & \cellcolor{blue!25}11.70\\
\bottomrule
\end{tabular}}
\end{table*}

The main reason for the worse performance of transformer-based architectures is the activation binarization in the attention mechanism. Compared to CNNs/MLPs that mainly use convolution or linear operations on the input data, the transformer-based architectures heavily rely on the attention mechanism, which involves multiplications operation between two binarized activations (between the query and key, attention possibilities, and value) and causes twice as much loss of information in one computation. Moreover, the binarization of activations is determined and cannot be adjusted during training, while the binarized weight can be learned through backward propagation during training. Since the attention mechanism requires more binarization of activation, the training of binarized transformers is more difficult and the inference is more unstable compared to CNNs/MLPs, while the binarized weights of the latter participate in each binarization computation and can be continuously optimized during training. We present more detailed discussions and empirical results in Appendix~\ref{subsec:DiscussionofBinarizedArchitectures}.

These results indicate that transformer-based architectures, with their unique attention mechanisms, require specialized binarization designs rather than direct binarization.
And the overall winner on the architecture track is the FDA algorithm, which performs best on both CNNs and transformers. The evaluation of these two tracks shows that binarization algorithms that use statistical channel-wise scaling factors and custom gradient approximation, such as FDA and ReActNet, have some degree of stability advantage.

\subsubsection{Corruption Robustness: Binarization Exhibits Robust to Corruption}

\textbf{Binarized networks can approach full-precision level robustness for corruption}. 
Interestingly, binarized networks demonstrate robustness comparable to full-precision counterparts when evaluated for corruption. Evaluation results on the CIFAR10-C dataset reveal that binarized networks perform similarly to full-precision networks on typical 2D image corruption tasks. In some cases, such as ReCU and XNOR-Net, binarized networks outperform their full-precision counterparts. If the same level of robustness to corruption is required, the binarized network version typically requires little additional design or supervision to achieve it. As such, binarized networks generally exhibit similar robustness to corruption as full-precision networks, which appears to be a general property of binarized networks rather than a specific characteristic of certain algorithms.
And our results also suggest that the reason binarized models are robust to data corruption cannot be directly attributed to the smaller model scale, but is more likely to be a unique characteristic of some binarization algorithms (Appendix~\ref{subsec:DiscussionofBinarization}).

We also analyze that the robustness of BNN to data corruption originates from the high discretization of parameters from the perspective of interpretability. Previous studies pointed out that the robustness of the quantization network is related to both the quantization bucket (range) and the magnitude of the noise~\cite{lin2018defensive}, where the former depends on the bit-width and quantizer and the latter depends on the input noise. Specifically, when the magnitude of the noise is small, the quantization bucket is capable of reducing the errors by eliminating small perturbations; however, when the magnitude of perturbation is larger than a certain threshold, quantization instead amplifies the errors. This fact precisely leads to the natural advantages of the binary network in the presence of natural noise. (1) 1-bit binarization enjoys the largest quantization bucket among all bits quantization. In binarization, the application of the sign function allows the binarizer to be regarded as consisting of a semi-open closed interval $[0, +\infty)$ and an open interval $(-\infty, 0)$ quantization bucket. So binarization is with the largest quantization bucket among all bit-width quantization (most of their quantization buckets are with closed intervals with limited range) and brings the largest noise tolerance to BNNs. (2) Natural data corruption (we evaluated in BiBench) is typically considered to have a smaller magnitude than adversarial noise which is always considered worst-case~\cite{ren2022benchmarking}. (1) and (2) show that when BNNs encounter corrupted inputs, the high discretization of parameters makes them more robust.

\begin{figure*}[t]
\centering
\vspace{-0.0in}
\includegraphics[width=\linewidth]{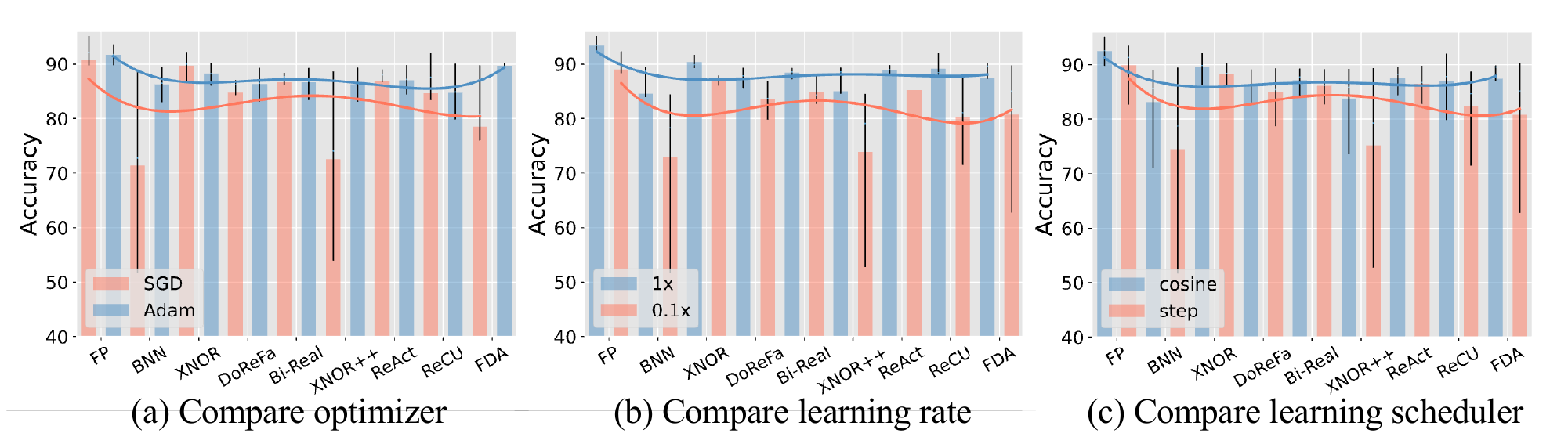}
\vspace{-0.25in}
\caption{Comparisons of accuracy under different training settings. }
\label{fig:ablation}
\end{figure*}

\subsection{Efficiency Tracks}
We analyze the efficiency metrics of training consumption, theoretical complexity, and hardware inference (Table~\ref{tab:eff}).

\subsubsection{Training Consumption: Binarization Could Be Stable yet Generally Expensive}
\label{sec:consumption}
We thoroughly examine the training cost of binarization algorithms on ResNet18 for CIFAR10 and present the sensitivity and training time results for different binarization algorithms in Table~\ref{tab:eff} and Figure~\ref{fig:speed_compute}, respectively.

\textbf{``Binarization$\not=$sensitivity": existing techniques can stabilize binarization-aware training}. It is commonly believed that the training of binarized networks is more sensitive to training settings than full-precision networks due to the representation limitations and gradient approximation errors introduced by the high degree of discretization. However, we find that the hyperparameter sensitivities of existing binarization algorithms are polarized, with some being even more hyperparameter-stable than the training of full-precision networks, while others fluctuate greatly. This variation is due to the different techniques used by the binarized operators of these algorithms. Hyperparameter-stable binarization algorithms often share the following characteristics: (1) \textit{channel-wise scaling factors} based on learning or statistics; (2) \textit{soft approximation} to reduce gradient error. These stable algorithms may not necessarily outperform others, but they can simplify the tuning process in production and provide reliable accuracy with a single training.

The preference for hyperparameter settings is also clear and stable binarized networks. Statistical results in Figure~\ref{fig:ablation} show that training with Adam optimizer,  the learning rate equaling to the full-precision network (1$\times$), and the CosineAnnealingLR scheduler is more stable than other settings. Based on this, we use this setting as part of the standard training pipelines when evaluating binarization.

\textbf{Soft approximation in binarization leads to a significant increase in training time}. Comparing the time consumed by each binarization algorithm, we found that the training time of algorithms using custom gradient approximation techniques such as Bi-Real and ReActNet increased significantly. The metric about the training time of FDA is even as high as 20.62\%, meaning it is almost 5$\times$ the training time of a full-precision network.

\subsubsection{Theoretical Complexity: Different Algorithms have Similar Complexity}

\textbf{There is a minor difference in theoretical complexity among binarization algorithms}. 
The leading cause of the difference in compression rate is each model's definition of the static scaling factor. For example, BNN does not use any factors and has the highest compression. In terms of theoretical acceleration, the main difference comes from two factors: the reduction in the static scaling factor also improves theoretical speedup, and real-time re-scaling and mean-shifting for activation add additional computation, such as in the case of ReActNet, which reduces the speedup by 0.11 times. In general, the theoretical complexity of each method is similar, with overall metrics in the range of $[12.71, 12.94]$. These results suggest that binarization algorithms should have similar inference efficiency.

\begin{table*}[t]
\centering
\vspace{-0.05in}
\caption{Deployment capability of different inference libraries on real hardware.}
\label{tab:cap}
\vspace{-0.15in}
\setlength{\tabcolsep}{3.5mm}
{
\begin{tabular}{lllcccc}
\toprule
Infer. Lib.      & Provider   & $s$ Granularity & $s$ Form & Flod BN & Act. Re-scaling & Act. Mean-shifting \\
\midrule
Larq                   & Larq       & Channel-wise          & FP32   & $\surd$       & $\times$                     & $\surd$                        \\
daBNN                  & JD         & Channel-wise          & FP32   & $\surd$       & $\times$                     & $\times$                        \\
\midrule
Algorithm & Deployable & $s$ Granularity & $s$ Form & Flod BN & Act. Re-scaling & Act. Mean-shifting \\
\midrule
BNN                    & {$\surd$}          & N/A                  & N/A      & N/A       & $\times$                     & $\times$                        \\
XNOR                   & {$\times$}          & Channel-wise          & FP32   & $\surd$       & $\surd$                     & $\times$                        \\
DoReFa                 & {$\surd$}          & Channel-wise          & FP32   & $\surd$       & $\times$                     & $\times$                        \\
Bi-Real                & {$\surd$}          & Channel-wise          & FP32   & $\surd$       & $\times$                     & $\times$                        \\
XNOR++                 & {$\times$}          & Spatial-wise          & FP32   & $\times$       & $\times$                     & $\times$                        \\
ReActNet               & {$\surd$}          & Channel-wise          & FP32   & $\surd$       & $\times$                     & $\surd$                        \\
ReCU                   & {$\surd$}          & Channel-wise          & FP32   & $\surd$       & $\times$                     & $\times$                        \\
FDA                    & {$\surd$}          & Channel-wise          & FP32   & $\surd$       & $\times$                     & $\times$                        \\
\bottomrule
\end{tabular}}
\end{table*}

\subsubsection{Hardware Inference: Immense Potential on Edge Devices Despite Limited Supports}
One of the advantages of the hardware inference track is that it provides valuable insights into the practicalities of deploying binarization in real-world settings. This track stands out from other tracks in this regard.

\textbf{Limited inference libraries lead to almost fixed paradigms of binarization deployment.} 
The availability of open-source inference libraries that support the deployment of binarization algorithms on hardware is quite limited. After investigating the existing options, we found that only Larq~\citep{geiger2020larq} and daBNN~\citep{zhang2019dabnn} offer complete deployment pipelines and primarily support deployment on ARM devices. As shown in Table~\ref{tab:cap}, both libraries support channel-wise scaling factors in the floating-point form that must be fused into the Batch Normalization (BN) layer. However, neither of them supports dynamic activation statistics or re-scaling during inference. Larq also includes support for mean-shifting activation with a fixed bias. These limitations in the available inference libraries' deployment capabilities have significantly impacted the practicality of deploying binarization algorithms. For example, the scale factor shape of XNOR++ caused its deployment to fail and XNOR also failed due to its activation re-scaling technique; BNN and DoReFa have different theoretical complexities but have the exact same true computation efficiency when deployed. These constraints have resulted in a situation where the vast majority of binarization methods have almost identical inference performance, with the mean-shifting operation of ReActNet on activation having only a slight impact on efficiency. As a result, binarized models must adhere to fixed deployment paradigms and have almost identical efficiency performance.

\begin{figure}
  \vspace{-0.1in}
  \begin{center}
    \includegraphics[width=1.\linewidth]{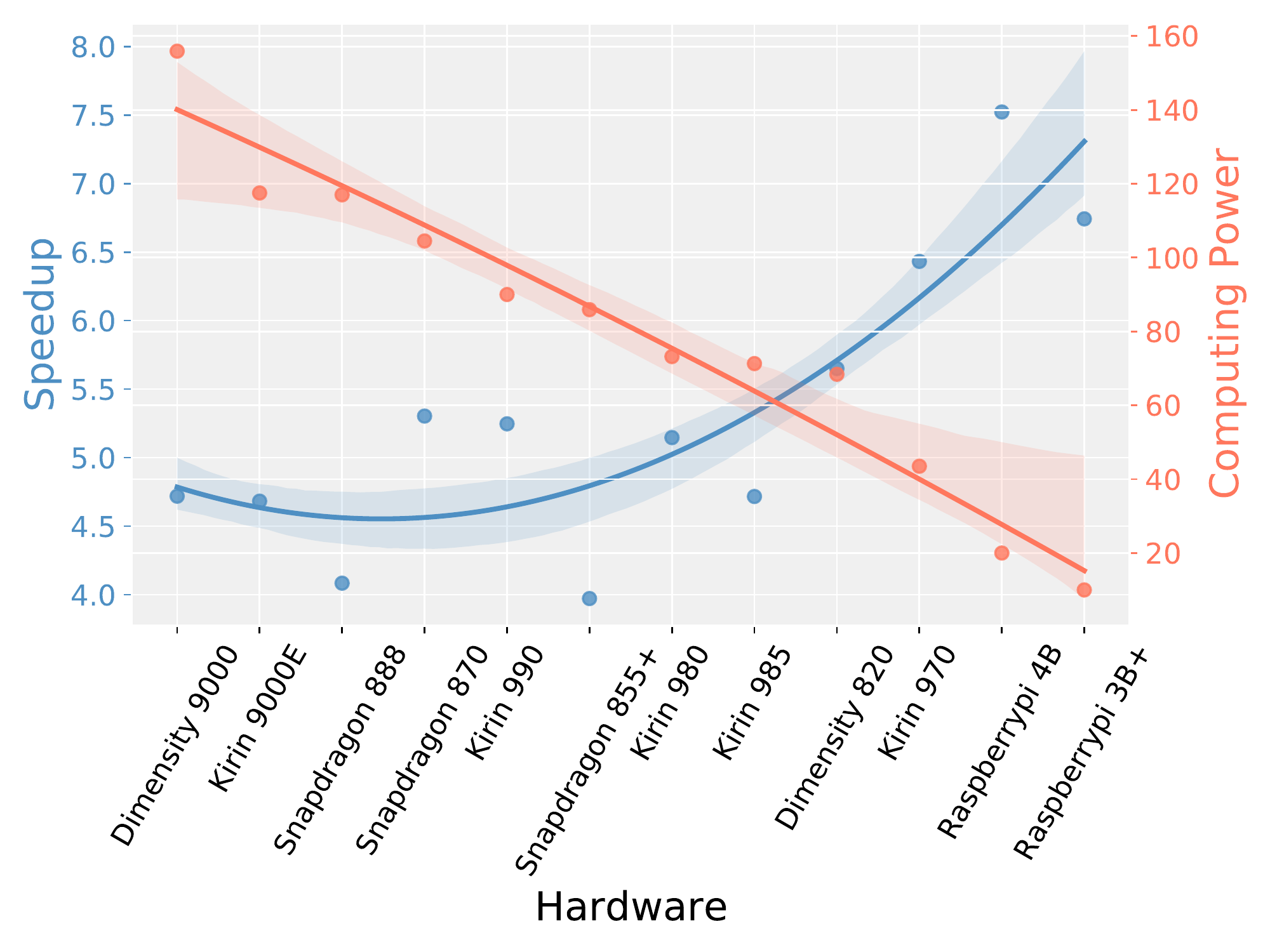}
  \end{center}
  \vspace{-0.2in}
  \caption{The lower the chip's computing power, the higher the inference speedup of deployed binarized models.}
  \label{fig:speed_compute}
\end{figure}

\textbf{Born for the edge: more promising for lower-power edge computing}.
After evaluating the performance of binarized models on a range of different chips, we found that the average speedup of the binarization algorithm was higher on chips with lower computing power (Figure~\ref{fig:speed_compute}). This counter-intuitive result is likely since higher-performance chips tend to have more acceleration from multi-threading when running floating-point models, leading to a relatively slower speedup of binarized models on these chips. In contrast, binarization technology is particularly effective on edge chips with lower performance and cost. Its extreme compression and acceleration capabilities can enable the deployment of advanced neural networks at the edge. These findings suggest that binarization is well-suited for low-power, cost-sensitive edge devices.

\subsection{Suggested Paradigm of Binarization Algorithm}
\label{subsec:SuggestedParadigm}
Based on our evaluation and analysis, we propose the following paradigm for achieving accurate and efficient network binarization using existing techniques:
(1) \textbf{Soft gradient approximation} presents great potential. It improves performance by increasing training rather than deployment cost. And all accuracy-winning algorithms adopt this technique, \textit{i.e.}, ReActNet, ReCU, and FDA. By further exploring this technique, FDA outperforms previous algorithms on the architecture track, indicating the great potential of the soft gradient approximation technique.
(2) \textbf{Channel-wise scaling factors} are currently the optimal option for binarization. The gain from the floating-point scaling factor is demonstrated in accuracy tracks, and deployable consideration limits its form to channel-wise. This means a balanced trade-off between accuracy and efficiency.
(3) \textbf{Pre-binarization parameter redistributing} is an optional but beneficial operation that can be implemented as a mean-shifting for weights or activations before binarization. Our findings indicate that this technique can significantly enhance accuracy with little added inference cost, as seen in ReActNet and ReCU.

It is important to note that, despite the insights gained from benchmarking on evaluation tracks, \textbf{none of the binarization techniques or algorithms work well across all scenarios so far}. Further research is needed to overcome the current limitations and mutual restrictions between production and deployment, and to develop binarization algorithms that consider both deployability and efficiency. Additionally, it would be helpful for inference libraries to support more advanced binarized operators. In the future, the focus of binarization research should be on addressing these issues.

\section{Discussion}
In this paper, we propose BiBench, a versatile and comprehensive benchmark toward the fundamentals of network binarization. BiBench covers 8 network binarization algorithms, 9 deep learning datasets (including a corruption one), 13 different neural architectures, 2 deployment libraries, 14 real-world hardware, and various hyperparameter settings. Based on these scopes, we develop evaluation tracks to measure the accuracy under multiple conditions and efficiency when deployed on actual hardware. By benchmark results and analysis, BiBench summarizes an empirically optimized paradigm with several critical considerations for designing accurate and efficient binarization algorithms. BiBench aims to provide a comprehensive and unbiased resource for researchers and practitioners working in model binarization. We hope BiBench can facilitate a fair comparison of algorithms through a systematic investigation with metrics that reflect the fundamental requirements for model binarization and serve as a foundation for applying this technology in broader and more practical scenarios. 

\noindent\textbf{Acknowledgement}. Supported by the National Key Research and Development Plan of China (2021ZD0110503), the National Natural Science Foundation of China (No. 62022009), the State Key Laboratory of Software Development Environment (SKLSDE-2022ZX-23), NTU NAP, MOE AcRF Tier 2 (T2EP20221-0033), and under the RIE2020 Industry Alignment Fund - Industry Collaboration Projects (IAF-ICP) Funding Initiative, as well as cash and in-kind contribution from the industry partner(s).

\nocite{langley00}

\newpage

\bibliography{example_paper}
\bibliographystyle{icml2023}

\newpage
\appendix
\onecolumn
{\LARGE\sc {Appendix for BiBench}\par}

\section{Details and Discussions of Benchmark}

\subsection{Details of Binarization Algorithm}
\label{app:algo}

Model compression methods including quantization~\cite{xiao2023benchmarking,liu2021ANP,guo2020multi,qin2022distribution,guo2020channel,qin2023diverse,chen2022empirical,qin2021hardware,zhang2021diversifying,guo2023cbanet,liu2022apsnet,Liu2019Perceptual} have been widely used in various deep learning fields~\cite{wang2021adversarial,DBLP:conf/ijcai/ZhaoXZLZL20,DBLP:journals/corr/abs-2303-06840,DBLP:journals/sigpro/ZhaoXZLZ20,guo2023towards,DBLP:journals/corr/abs-2303-08942,guo2021jointpruning,Liu2023Xadv}, including computer vision~\cite{guo2020model,wang2021universal,ding2022towards,DBLP:conf/cvpr/ZhaoZXLP22,wang2021dual,DBLP:journals/corr/abs-2211-14461,wang2022defensive,yin2021improving,DBLP:journals/tcsv/ZhaoXZLZL22}, language understanding~\cite{bai2021binarybert,wang2022generating}, speech recognition~\cite{qin2023bifsmnv2}, etc.
Previous research has deemed lower bit-width quantization methods as more aggressive~\citep{rusci2020memory,choukroun2019low,ijcai2022p603}, as they often provide higher compression and faster processing at the cost of lower accuracy. Among all quantization techniques, 1-bit quantization (binarization) is considered the most aggressive~\citep{ijcai2022p603}, as it poses significant accuracy challenges but offers the greatest compression and speed benefits.

\textit{Training}. When training a binarized model, the sign function is commonly used in the forward pass, and gradient approximations such as STE are applied during the backward pass to enable the model to be trained. Since the parameters are quantized to binary, network binarization methods typically use a simple sign function as the quantizer rather than using the same quantizer as in multi-bit (2-8 bit) quantization \citep{Gong:iccv19,gholamisurvey}. Specifically, as \cite{Gong:iccv19} describes, for multi-bit uniform quantization, given the bit width b and the floating-point activation/weight $x$ following in the range $(l, u)$, the complete quantization-dequantization process of uniform quantization can be defined as

\begin{equation}
Q_U(\boldsymbol{x})=\operatorname{round}\left(\frac{\boldsymbol{x}}{\Delta}\right) \Delta,
\end{equation}

where the original range $(l, u)$ is divided into $2^b-1$ intervals Pi, $i \in (0, 1, \cdots, 2^b-1)$, and $\Delta = \frac{u-l}{2^b-1}$ is the interval length. When $b=1$, the $Q_U (\boldsymbol{x})$ equals the sign function, and the binary function is expressed as

\begin{equation}
Q_B(\boldsymbol{x}) = \operatorname{sign}(\boldsymbol{x}).
\end{equation}

Therefore, binarization can be regarded formally as the 1-bit specialization of quantization.

\textit{Deployment}. 
For efficient deployment on real-world hardware, binarized parameters are grouped in blocks of 32 and processed simultaneously using 32-bit instructions, which is the key principle for achieving acceleration. To compress binary algorithms, instructions such as XNOR (or the combination of EOR and NOT) and popcount are used to enable the deployment of binarized networks on real-world hardware. The XNOR (exclusive-XOR) gate is a combination of an XOR gate and an inverter, and XOR (also known as EOR) is a common instruction that has long been available in assembly instructions for all target platforms. The popcount instruction, or Population Count per byte, counts the number of bits with a specific value in each vector element in the source register, stores the result in a vector and writes it to the destination register~\citep{arm64}. This instruction is used to accelerate the inference of binarized networks~\citep{BNN,rastegari2016xnor} and is widely supported by various hardware, such as the definitions of popcount in ARM and x86 in \citep{arm64} and \citep{x86}, respectively.

\textit{Comparison with other compression techniques}.
Current network compression technologies primarily focus on reducing the size and computation of full-precision models. Knowledge distillation, for instance, guides the training of small (student) models using the intermediate features and/or soft outputs of large (teacher) models~\citep{hinton2015distilling,xu2018training,chen2018darkrank,Yim2017A,zagoruyko2017paying}. Model pruning~\citep{han2015learning, han2016deep, he2017channel} and low-rank decomposition~\citep{denton2014exploiting,lebedev2015speeding,jaderberg2014speeding,lebedev2016fast} also reduce network parameters and computation via pruning and low-rank approximation. Compact model design, on the other hand, creates a compact model from scratch \citep{mobilenet,mobilenet_v2,shufflenet,shufflenet_v2}. While these compression techniques effectively decrease the number of parameters, the compressed models still use 32-bit floating-point numbers, which leaves scope for additional compression using model quantization/binarization techniques. Compared to multi-bit (2-8 bit) model quantization, which compresses parameters to integers\citep{gong2014compressing,wu2016quantized,Vanhoucke2011ImprovingTS,gupta2015deep}, binarization directly applies the sign function to compress the model to a more compact 1-bit~\citep{rusci2020memory,choukroun2019low,ijcai2022p603,shang2022network,qin2020forward}. Additionally, due to the use of binary parameters, bitwise operations (XNOR and popcount) can be applied during inference at deployment instead of integer multiply-add operations in 2-8 bit model quantization. As a result, binarization is considered to be more hardware-efficient and can achieve greater speedup than multi-bit quantization.

\begin{figure}[t]
\centering
\includegraphics[width=0.9\linewidth]{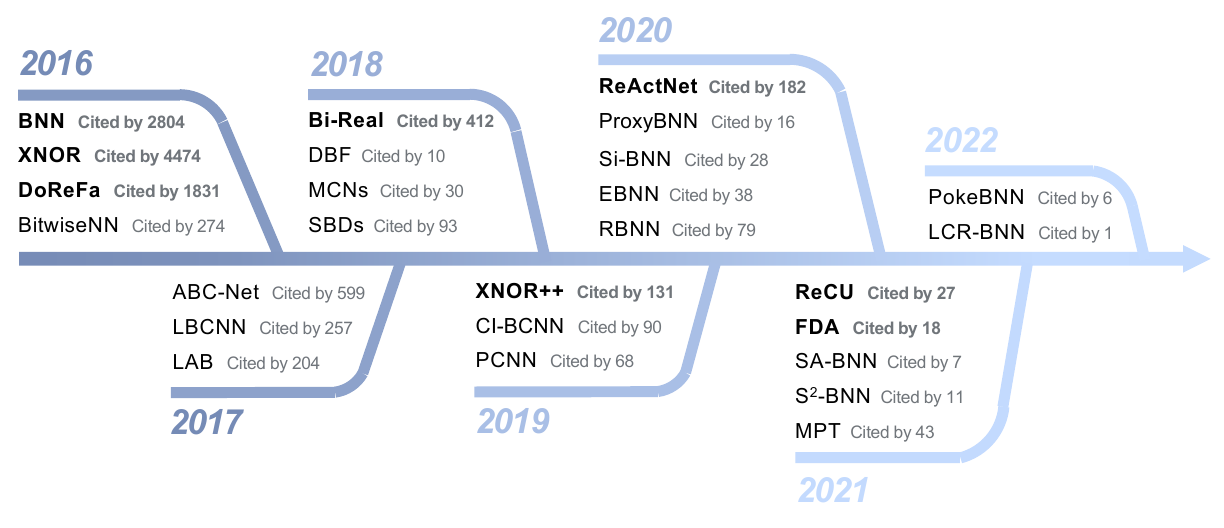}
\vspace{-0.15in}
\caption{Timeline of the operator-level binarization algorithms we have considered. The algorithms selected for BiBench are in bold, and the citation is counted till 2023/01/25.}
\label{fig:timeline}
\end{figure}

\textbf{Selection Rules}:

When creating the BiBench, we considered various binarization algorithms with enhanced operator techniques in binarization research, and the timeline of considered algorithms is Figure~\ref{fig:timeline} and we list their details in Table~\ref{tab:all_algos}.
We follow two general rules when selecting algorithms for our benchmark:

(1) \textbf{The selected algorithms should function on binarized operators} that are the fundamental components for binarized networks (as discussed in Section~\ref{subsec:NetworkBinarization}). 
We exclude algorithms and techniques that require specific local structures or training pipelines to ensure a fair comparison.

(2) \textbf{The selected algorithms should have an extensive influence to be representative}, \textit{i.e.}, selected from widely adopted algorithms or the most advanced ones.

Specifically, we chose algorithms based on the following detailed criteria to ensure representativeness and fairness in evaluations: Operator Techniques (Yes/No), Year, Conference, Citations (up to 2023/01/25), Open source availability (Yes/No), and Specific Structure / Training-pipeline requirements (Yes/No/Optional). 

We analyze the techniques proposed in these works. Following the general rules we mentioned, all considered binarization algorithms should have significant contributions to the improvement of the binarization operator (Operator Techniques: Yes) and should not include techniques that are bound to specific architectures and training pipelines to complete well all the evaluations of the learning task, neural architecture, and training consumption tracks in BiBench (Specified Structure / Training-pipeline: No/Optional, Optional means the techniques are included but can be decoupled with binarized operator totally). 
We also consider the impact and reproducibility of these works. We prioritized the selection of works with more than 100 citations, which means they are more discussed and compared in binarization research and thus have higher impacts. Works in 2021 and later are regarded as the SOTA binarization algorithms and prioritized. Furthermore, we hope the selected works have official open-source implementations for reproducibility.

Based on the above selections, eight binarization algorithms, \textit{i.e.}, BNN, XNOR-Net, DoReFa-Net, Bi-Real Net, XNOR-Net++, ReActNet, FDA, and ReCU, stand out and are fully evaluated by our BiBench.

\begin{table}[t]
\caption{The considered operator-level binarization algorithms and our final selections in BiBench. Bold means that the algorithm has an advantage in that column.}
\label{tab:all_algos}
\vspace{-0.1in}
\setlength{\tabcolsep}{1.4mm}{
\begin{tabular}{llccccc}
\toprule
{Algorithm} & {Year} & {Conference} & {\tabincell{c}{Citation\\(2023/01/25)}} & {\tabincell{c}{Operator\\Techniques}} & {\tabincell{c}{Open\\Source}} & {\tabincell{c}{Specified Structure /\\Training-pipeline}} \\
\midrule
{BitwiseNN}~\citep{kim2016bitwise}   & {2016}       & ICMLW        & \textbf{274}    & \textbf{Yes}     & No    & \textbf{No}       \\
\rowcolor{blue!10}\textbf{DoReFa}~\citep{Dorefa-Net}       & 2016     & ArXiv          & \textbf{1831}     & \textbf{Yes}     & \textbf{Yes}    & \textbf{No}       \\
\rowcolor{blue!10}\textbf{XNOR-Net}~\citep{rastegari2016xnor}     & 2016     & ECCV           & \textbf{4474}    & \textbf{Yes}     & \textbf{Yes}    & \textbf{No}       \\
\rowcolor{blue!10}\textbf{BNN}~\citep{BinaryNet}   & {2016}       & NeurIPS        & \textbf{2804}    & \textbf{Yes}     & \textbf{Yes}    & \textbf{No}       \\
{LBCNN}~\citep{juefei2017local}       & 2017     & CVPR          & \textbf{257}     & \textbf{Yes}     & \textbf{Yes}    & Yes       \\
{LAB}~\citep{hou2017loss}       & 2017     & ICLR          & \textbf{204}     & \textbf{Yes}     & \textbf{Yes}    & Yes       \\
{ABC-Net}~\citep{ABC-Net}       & 2017     & NeurIPS          & \textbf{599}     & \textbf{Yes}     & \textbf{Yes}    & Yes       \\
{DBF}~\citep{tseng2018deterministic}         & 2018     & IJCAI           & 10     & \textbf{Yes}     & No    & Yes       \\
{MCNs}~\citep{wang2018modulated}         & 2018     & CVPR           & 30     & \textbf{Yes}     & No    & Yes        \\
{SBDs}~\citep{hu2018training}         & 2018     & ECCV           & 93     & \textbf{Yes}     & No    & \textbf{No}        \\
\rowcolor{blue!10}\textbf{Bi-Real Net}~\citep{Bi-Real}         & 2018     & ECCV           & \textbf{412}     & \textbf{Yes}     & \textbf{Yes}    & \textbf{Opt}        \\
PCNN~\citep{gu2019projection}        & {2019}        & AAAI           & 68        & \textbf{Yes}     & No    & Yes        \\
CI-BCNN~\citep{wang2019learning}        & {2019}        & CVPR           & 90        & \textbf{Yes}     & \textbf{Yes}    & Yes        \\
\rowcolor{blue!10}\textbf{XNOR-Net++}~\citep{bulat2019xnor}          & 2019     & BMVC           & \textbf{131}     & \textbf{Yes}     & \textbf{Yes}    & \textbf{No}       \\
ProxyBNN~\citep{he2020proxybnn}         & 2020     & ECCV           & 16        & \textbf{Yes}     & No       & Yes       \\
Si-BNN~\citep{wang2020sparsity}         & 2020     & AAAI           & 28        & \textbf{Yes}     & No       & \textbf{No}       \\
EBNN~\citep{bulat2020high}    & 2020          & ICLR           & 38        & \textbf{Yes}     & \textbf{Yes}    & Yes       \\
RBNN~\citep{lin2020rotated}           & 2020     & NeurIPS        & 79        & \textbf{Yes}     & \textbf{Yes}    & \textbf{No}       \\
\rowcolor{blue!10}\textbf{ReActNet}~\citep{liu2020reactnet}     & 2020     & ECCV           & \textbf{182}     & \textbf{Yes}     & \textbf{Yes}    & \textbf{Opt}        \\
SA-BNN~\citep{liu2021sa}    & \textbf{2021}          & AAAI           & 7        & \textbf{Yes}     & No    & \textbf{No}       \\
S$^2$-BNN~\citep{shen2021s2}    & \textbf{2021}          & CVPR           & 11        & \textbf{Yes}     & \textbf{Yes}    & {Yes}       \\
MPT~\citep{diffenderfer2021multi}    & \textbf{2021}          & ICLR           & 43      & \textbf{Yes}     & \textbf{Yes}    & Yes       \\
\rowcolor{blue!10}\textbf{FDA}~\citep{xu2021learning}   & \tabincell{l}{\textbf{2021}}        & NeurIPS        & 18   
& \textbf{Yes}     & \textbf{Yes}    & \textbf{No}       \\
\rowcolor{blue!10}\textbf{ReCU}~\citep{xu2021recu}    & \textbf{2021}          & ICCV           & 27        & \textbf{Yes}     & \textbf{Yes}    & \textbf{No}       \\
LCR-BNN~\citep{shang2022lipschitz}     & \textbf{2022} & ECCV    & {1}       & \textbf{Yes}     & \textbf{Yes}         & Yes        \\ 
PokeBNN~\citep{zhang2022pokebnn}    & \textbf{2022}          & CVPR           & 6        & \textbf{Yes}     & \textbf{Yes}    & {Yes}       \\
\bottomrule
\end{tabular}}
\end{table}

\textbf{Algorithm Details}:

\textbf{BNN}~\citep{courbariaux2016binarized}:
During the training process, BNN uses the straight-through estimator (STE) to calculate gradient $\boldsymbol{g_{x}}$ which takes into account the saturation effect:

\begin{equation}
\label{eq:app_ste}
\mathtt{sign}(\boldsymbol{x})=
\begin{cases}
+1,& \mathrm{if} \ \boldsymbol x \ge 0\\
-1,& \mathrm{otherwise}
\end{cases}\qquad
\boldsymbol{g_{x}}=
\begin{cases}
\boldsymbol{g_b},& \mathrm{if} \ \boldsymbol x \in \left(-1, 1\right)\\
0,& \mathrm{otherwise}.
\end{cases}
\end{equation}

And during inference, the computation process is expressed as

\begin{equation}
\boldsymbol o = \operatorname{sign}(\boldsymbol{a}) \circledast \operatorname{sign}(\boldsymbol{w}),
\end{equation}

where $\circledast$ indicates a convolutional operation using XNOR and bitcount operations.

\textbf{XNOR-Net}~\citep{rastegari2016xnor}:
XNOR-Net obtains the channel-wise scaling factors $\boldsymbol \alpha=\frac{\left\|\boldsymbol{w}\right\|}{\left|\boldsymbol{w}\right|}$ for the weight and $\boldsymbol{K}$ contains scaling factors $\beta$ for all sub-tensors in activation $\boldsymbol{a}$. We can approximate the convolution between activation $\boldsymbol{a}$ and weight $\boldsymbol{w}$ mainly using binary operations:

\begin{equation}
\label{eq:xnor-net}
\boldsymbol o = (\operatorname{sign}(\boldsymbol{a}) \circledast \operatorname{sign}(\boldsymbol{w})) \odot \boldsymbol{K} \boldsymbol \alpha,
\end{equation}

where $\boldsymbol{w} \in \mathbb{R}^{c \times w \times h}$ and $\boldsymbol{a} \in \mathbb{R}^{c \times w_{\text {in }} \times h_{\text {in }}}$ denote the weight and input tensor, respectively. And the STE is also applied in the backward propagation of the training process.

\textbf{DoReFa-Net}~\citep{Dorefa-Net}:
DoReFa-Net applies the following function for $1$-bit weight and activation:

\begin{equation}
\label{eq:dorefa}
\boldsymbol o = (\operatorname{sign}(\boldsymbol{a}) \circledast \operatorname{sign}(\boldsymbol{w})) \odot \boldsymbol \alpha.
\end{equation}

And the STE is also applied in the backward propagation with the full-precision gradient.

\textbf{Bi-Real Net}~\citep{liu2018bi}:
Bi-Real Net proposes a piece-wise polynomial function as the gradient approximation function:

\begin{equation}
\operatorname{bireal}\left(\boldsymbol{a}\right)=\left\{\begin{array}{lr}
-1 & \text { if } \boldsymbol{a}<-1 \\
2 \boldsymbol{a}+\boldsymbol{a}^2 & \text { if }-1 \leqslant \boldsymbol{a}<0 \\
2 \boldsymbol{a}-\boldsymbol{a}^2 & \text { if } 0 \leqslant \boldsymbol{a}<1 \\
1 & \text { otherwise }
\end{array}, \quad \frac{\partial \operatorname{bireal}\left(\boldsymbol{a}\right)}{\partial \boldsymbol{a}}= \begin{cases}2+2 \boldsymbol{a} & \text { if }-1 \leqslant \boldsymbol{a}<0 \\
2-2 \boldsymbol{a} & \text { if } 0 \leqslant \boldsymbol{a}<1 \\
0 & \text { otherwise }\end{cases}\right. .
\end{equation}

And the forward propagation of Bi-Real Net is the same as Eq.~(\ref{eq:dorefa}).

\textbf{XNOR-Net++}~\citep{bulat2019xnor}:
XNOR-Net++ proposes to re-formulate Eq.~(\ref{eq:xnor-net}) as:

\begin{equation}
\boldsymbol{o} = (\operatorname{sign}(\boldsymbol{a}) \circledast \operatorname{sign}(\boldsymbol{w})) \odot \boldsymbol \Gamma,
\end{equation}

and we adopt the $\boldsymbol \Gamma$ as the following form in experiments (achieve the best performance in the original paper):

\begin{equation}
\boldsymbol \Gamma=\boldsymbol \alpha \otimes \boldsymbol \beta \otimes \boldsymbol \gamma, \quad \boldsymbol \alpha \in \mathbb{R}^{\boldsymbol{o}}, \boldsymbol \beta \in \mathbb{R}^{h_{\text {out }}}, \boldsymbol \gamma \in \mathbb{R}^{w_{\text {out }}},
\end{equation}

where $\otimes$ denotes the outer product operation, and $\boldsymbol \alpha$, $\boldsymbol \beta$, and $\boldsymbol \gamma$ are learnable during training.


\textbf{ReActNet}~\citep{liu2020reactnet}:
ReActNet defines an RSign as a binarization function with channel-wise learnable thresholds:

\begin{equation}
\boldsymbol{x}=\operatorname{rsign}\left(\boldsymbol{x}\right)=\left\{\begin{array}{ll}
+1, & \text { if } \boldsymbol{x}>\boldsymbol \alpha \\
-1, & \text { if } \boldsymbol{x} \leq \boldsymbol \alpha
\end{array} .\right.
\end{equation}

where $\boldsymbol \alpha$ is a learnable coefficient controlling the threshold. And the forward propagation is

\begin{equation}
\boldsymbol o = (\operatorname{rsign}(\boldsymbol{a}) \circledast \operatorname{sign}(\boldsymbol{w})) \odot \boldsymbol \alpha.
\end{equation}

where $\boldsymbol \alpha$ denotes the channel-wise scaling factors of weights. Parameters in RSign are optimized end-to-end with other parameters in the network. The gradient of $\boldsymbol \tau$ in RSign can be simply derived by the chain rule as:

\begin{equation}
\frac{\partial \mathcal{L}}{\partial \boldsymbol \tau}= \frac{\partial \mathcal{L}}{\partial h\left(\boldsymbol x\right)} \frac{\partial h\left(\boldsymbol x\right)}{\partial \boldsymbol \tau},
\end{equation}

where $\mathcal{L}$ represents the loss function and $\frac{\partial \mathcal{L}}{\partial h\left(\boldsymbol x\right)}$ denotes the gradients from deeper layers. The derivative $\frac{\partial h\left(\boldsymbol x\right)}{\partial \boldsymbol \tau}$ can be easily computed as

\begin{equation}
\frac{\partial h\left(\boldsymbol x\right)}{\partial \boldsymbol \tau}=-1.
\end{equation}

The estimator of the sign function for activation is

\begin{equation}
\operatorname{react} (\boldsymbol{a}) =
\begin{cases}
-1 \  &\text{if} \ \boldsymbol{a} < -1; \\
\ 2\boldsymbol{a} + \boldsymbol{a}^2 \ &\text{if} \ -1 \leq \boldsymbol{a} < 0; \\
\  2\boldsymbol{a} - \boldsymbol{a}^2; \ &\text{if} \ 0 \leq \boldsymbol{a} < 1;\\
\ 1\  &\text{otherwise} ,
\end{cases}
\end{equation}

\begin{equation}
\frac{\partial \operatorname{react} (\boldsymbol{a})}{\partial \boldsymbol{a}} =
\begin{cases}
2 + 2\boldsymbol{a} \ &\text{if} \ - 1 \leq \boldsymbol{a} < 0; \\
2 - 2\boldsymbol{a}; \ &\text{if}\ 0 \leq \boldsymbol{a} < 1;\\
0 \ &\text{otherwise} .
\end{cases}
\end{equation}

And the estimator for weight is the STE form straightforwardly as Eq.~\eqref{eq:app_ste}.


\textbf{ReCU}~\citep{xu2021recu}:
As described in their paper, ReCU is formulated as

\begin{equation}
\operatorname{recu}(\boldsymbol{w})=\max \left(\min \left(\boldsymbol{w}, Q_{(\tau)}\right), Q_{(1-\tau)}\right),
\end{equation}

where $Q_{(\tau)}$ and $Q_{(1-\tau)}$ denote the $\tau$ quantile and $1-\tau$ quantile of $\boldsymbol{w}$, respectively. After applying balancing, weight standardization, and the ReCU to $\boldsymbol w$, the generalized probability density function of $\boldsymbol w$ can be written as follows

\begin{equation}
f(w) =   
\begin{cases}
\frac{1}{2b} 
\operatorname{exp}\left(\frac{-|\boldsymbol w|}{b}\right), \
&\text{if} \ |\boldsymbol w| < Q(\tau) ; \\
1 - \tau, \ 
&\text{if} \ |\boldsymbol w| = Q(\tau) ; \\
0, \ 
&\text{otherwise},
\end{cases}
\end{equation}

where $b$ is obtained via

\begin{equation}
b = \operatorname{mean}(|\boldsymbol {w}|),
\end{equation}

where $\operatorname{mean}(|\cdot|)$ returns the mean of the absolute values of the inputs. The gradient of the weights is obtained by applying the chain derivation rule to the above equation, where the estimator applied to the sign function is STE. As for the gradient w.r.t. the activations, ReCU considers the piecewise polynomial function as follows

\begin{equation}
\frac{\partial {L}}{\partial \operatorname{sign}(\boldsymbol{a})}
=\frac{\partial {L}}{\partial 
\operatorname{sign}(\boldsymbol{a})}
\cdot \frac{\partial \operatorname{sign}(\boldsymbol{a})}{\partial \boldsymbol{a}} 
\approx \frac{\partial {L}}{\partial 
\operatorname{sign}(\boldsymbol{a})} 
\cdot \frac{\partial F\left(\boldsymbol{a}\right)}{\partial \boldsymbol{a}},
\end{equation}

where 

\begin{equation}
\frac{\partial F\left(\boldsymbol{a}\right)}{\partial \boldsymbol{a}}=
\begin{cases}
2+2 \boldsymbol{a},  &\text { if }-1 \leq \boldsymbol{a}<0;\\
2-2 \boldsymbol{a}, &\text { if } 0 \leq \boldsymbol{a}<1;\\
0, &\text { otherwise }.
\end{cases}
\end{equation}

And other implementations also strictly follow the original paper and official code.

\textbf{FDA}~\citep{xu2021learning}:
FDA applies the following forward propagation in the operator:

\begin{equation}
\boldsymbol o = (\operatorname{rsign}(\boldsymbol{a}) \circledast \operatorname{sign}(\boldsymbol{w})) \odot \boldsymbol \alpha,
\end{equation}

and computes the gradients of $\boldsymbol o$ in the backward propagation as:

\begin{equation}
\frac{\partial \ell}{\partial \mathbf{t}}=\frac{\partial \ell}{\partial \boldsymbol{o}} \boldsymbol{w}_2^{\top} \odot\left(\left(\mathbf{t} \boldsymbol{w}_1\right) \geq 0\right) \boldsymbol{w}_1^{\top}
+\frac{\partial \ell}{\partial \boldsymbol{o}} \eta^{\prime}(\mathbf{t})
+\frac{\partial \ell}{\partial \boldsymbol{o}} \odot \frac{4 \omega}{\pi} \sum_i \cos ((2 i+1) \omega \mathbf{t}),
\end{equation}

where $\frac{\partial \ell}{\partial \boldsymbol{o}}$ is the gradient from the upper layers, $\odot$ represents element-wise multiplication, and $\frac{\partial \ell}{\partial \mathbf{t}}$ is the partial gradient on $\mathbf{t}$ that backward propagates to the former layer. And $\boldsymbol{w}_1$ and $\boldsymbol{w}_2$ are weights in the original models and the noise adaptation modules respectively. FDA updates them as

\begin{equation}
\frac{\partial \ell}{\partial \boldsymbol{w}_1}=\mathbf{t}^{\top} \frac{\partial \ell}{\partial \boldsymbol{o}} \boldsymbol{w}_2^{\top} \odot\left(\left(\mathbf{t} \boldsymbol{w}_1\right) \geq 0\right),\qquad
\frac{\partial \ell}{\partial \boldsymbol{w}_2}=\sigma\left(\mathbf{t} \boldsymbol{w}_1\right)^{\top} \frac{\partial \ell}{\partial \boldsymbol{o}}.
\end{equation}

In Table~\ref{tab:algos_detail}, we detail the techniques included in the selected binarization algorithms.

\begin{table}[!ht]
\centering
\caption{The details of the selected binarization algorithms in BiBench.}
\label{tab:algos_detail}
\vspace{-0.1in}
\setlength{\tabcolsep}{2.5mm}{
\begin{threeparttable}
\begin{tabular}{lcccccc}
\toprule
\multirow{2}{*}{Algorithm} & \multicolumn{2}{c}{Scaling Factor} & \multicolumn{2}{c}{Parameter Redistribution} & \multicolumn{2}{c}{Gradient Approximation} \\
~ & weight & activation & weight & activation & weight & activation \\ \midrule
BNN & w/o & w/o & w/o & w/o & STE & STE \\ \midrule
XNOR & \tabincell{c}{Statistics\\by Channel} & \tabincell{c}{Statistics\\by Channel} & w/o & w/o & STE & STE \\ \midrule
DoReFa & \tabincell{c}{Statistics\\by Layer} & w/o & w/o & w/o & STE & STE \\ \midrule
Bi-Real & \tabincell{c}{Statistics\\by Channel} & w/o & w/o & w/o & STE & \tabincell{c}{Differentiable\\Piecewise\\Polynomial\\Function} \\ \midrule
XNOR++ & \tabincell{c}{Learned by\\Custom-size\\($o\times h_\text{out} \times w_\text{out}$)} & w/o & w/o & w/o & STE & STE \\ \midrule
ReActNet & \tabincell{c}{Statistics\\by Channel} & w/o & w/o & w/o & STE & \tabincell{c}{Differentiable\\Piecewise\\Polynomial\\Function} \\ \midrule
ReCU & \tabincell{c}{Statistics\\by Channel} & w/o & \tabincell{c}{balancing\\(mean-shifting)} & w/o & \tabincell{c}{Rectified\\Clamp Unit} & \tabincell{c}{Rectified\\Clamp Unit} \\ \midrule
FDA & \tabincell{c}{Statistics\\by Channel} & w/o & w/o & mean-shifting & \tabincell{c}{Decomposing\\Sign with\\Fourier Series} & \tabincell{c}{Decomposing\\Sign with\\Fourier Series} \\ \bottomrule
\end{tabular}
\begin{tablenotes}
\item[1] ``STE" indicates the Straight Through Estimator, and ``w/o" means no special technique is used.
\end{tablenotes}
\end{threeparttable}}
\end{table}

\subsection{Details of Learning Tasks}
\label{app:task}

\textbf{Selection Rules}:

To evaluate the performance of the binarization algorithm in a wide range of learning tasks, we must select a variety of representative tasks. Firstly, we choose representative perception modalities in deep learning, including (2D/3D) vision, text, and speech. These modalities have seen rapid progress and broad impact, so we select specific tasks and datasets within these modalities.
Specifically, (1) in the 2D vision modality, we select the essential image classification task and the prevalent object detection task, with datasets including CIFAR10 and ImageNet for the former and Pascal VOC and COCO for the latter. ImageNet and COCO datasets are more challenging and significant, while CIFAR10 and Pascal VOC are more fundamental.
For other modalities, binarization is still challenging even with the underlying tasks and datasets in the field, since there are few related binarization studies: (2) in the 3D vision modality, the basic point cloud classification ModelNet40 dataset is selected to evaluate the binarization performance, which is regarded as one of the most fundamental tasks in 3D point cloud research and is widely studied. (3) In the text modality, the General Language Understanding Evaluation (GLUE) benchmark is usually recognized as the most popular dataset, including nine sentence- or sentence-pair language understanding tasks. (4) The keyword spotting task was chosen as the base task in the speech modality, specifically the Google Speech Commands classification dataset.

Based on the above reasons and rules, we have selected a series of challenging and representative tasks for BiBench to evaluate binarization comprehensively and have obtained a series of reliable and informative conclusions and experiences.

\textbf{Dataset Details:}

\textbf{CIFAR10}~\citep{CIFAR}:
The CIFAR-10 dataset (Canadian Institute For Advanced Research) is a collection of images commonly used to train machine learning and computer vision algorithms. This dataset is widely used for image classification tasks. There are 60,000 color images, each of which measures 32x32 pixels. All images are categorized into 10 different classes: airplanes, cars, birds, cats, deer, dogs, frogs, horses, ships, and trucks. Each class has 6000 images, where 5000 are for training and 1000 are for testing. 
The evaluation metric of the CIFAR-10 dataset is accuracy, defined as:
\begin{equation}
    Accuracy = \frac{TP+TN}{TP+TN+FP+FN},
\end{equation}
where \textit{TP} (True Positive) means cases correctly identified as positive, \textit{TN} (True Negative) means cases correctly identified as negative, \textit{FP} (False Positive) means cases incorrectly identified as positive and \textit{FN} (False Negative) means cases incorrectly identified as negative. To estimate the accuracy, we should calculate the proportion of \textit{TP} and \textit{TN} in all evaluated cases.

\textbf{ImageNet}~\citep{krizhevsky2012imagenet}:
ImageNet is a dataset of over 15 million labeled high-resolution images belonging to roughly 22,000 categories. 
The images are collected from the web and labeled by human labelers using a crowd-sourced image labeling service called Amazon Mechanical Turk. As part of the Pascal Visual Object Challenge, ImageNet Large-Scale Visual Recognition Challenge (ILSVRC) was established in 2010. There are approximately 1.2 million training images, 50,000 validation images, and 150,000 testing images in total in ILSVRC. ILSVRC uses a subset of ImageNet, with about 1000 images in each of the 1000 categories. 

ImageNet also uses accuracy to evaluate the predicted results, which is defined above.

\textbf{Pascal VOC07}~\citep{hoiem2009pascal}:
The PASCAL Visual Object Classes 2007 (VOC07)  dataset contains 20 object categories including vehicles, households, animals, and other: airplane, bicycle, boat, bus, car, motorbike, train, bottle, chair, dining table, potted plant, sofa, TV/monitor, bird, cat, cow, dog, horse, sheep, and person. 
As a benchmark for object detection, semantic segmentation, and object classification, this dataset contains pixel-level segmentation annotations, bounding box annotations, and object class annotations. 
The VOC07 dataset uses mean average precision($mAP$) to evaluate results, which is defined as:
\begin{equation}
    mAP = \frac{1}{n} \sum_{k=1}^{k=n} AP_k
\end{equation}
where $AP_k$ denotes the average precision of the k-th category, which calculates the area under the precision-recall curve:
\begin{equation}
    AP_k = \int^1_0 p_k(r)dr.
\end{equation}
Especially for VOC07, we apply 11-point interpolated $AP$, which divides the recall value to $\{ 0.0, 0.1, \dots, 1.0 \}$  and then computes the average of maximum precision value for these 11 recall values as:
\begin{align}
    AP = \frac{1}{11} \sum_{r \in \{ 0.0, \dots, 1.0 \}} AP_r = \frac{1}{11} \sum_{r \in \{ 0.0, \dots, 1.0 \} } p_{\textrm{interp}}r.
\end{align}
The maximum precision value equals to the right of its recall level:
\begin{equation}
    p_{\textrm{interp}}r = \max_{\tilde{r} \geq r} p(\tilde{r}). 
\end{equation}

\textbf{COCO17}~\citep{lin2014microsoft}:
The MS COCO (Microsoft Common Objects in Context) dataset is a large-scale object detection, segmentation, key-point detection, and captioning dataset. The dataset consists of 328K images. 
According to community feedback, in the 2017 release, the training/validation split was changed from 83K/41K to 118K/5K. And the images and annotations are the same. The 2017 test set is a subset of 41K images from the 2015 test set. Additionally, 123K images are included in the unannotated dataset. 
The COCO17 dataset also uses mean average precision ($mAP$) as defined above PASCAL VOC07 uses, which is defined as above.


\textbf{ModelNet40}~\citep{wu20153d}:
The ModelNet40 dataset contains point clouds of synthetic objects. As the most widely used benchmark for point cloud analysis, ModelNet40 is popular due to the diversity of categories, clean shapes, and well-constructed dataset. 
In the original ModelNet40, 12,311 CAD-generated meshes are divided into 40 categories, where 9,843 are for training, and 2,468 are for testing. The point cloud data points are sampled by a uniform sampling method from mesh surfaces and then scaled into a unit sphere by moving to the origin. 
The ModelNet40 dataset also uses accuracy as the metric, which has been defined above in CIFAR10. 

\textbf{ShapeNet}~\citep{chang2015shapenet}:
ShapeNet is a large-scale repository for 3D CAD models developed by researchers from Stanford University, Princeton University, and the Toyota Technological Institute in Chicago, USA. 

Using WordNet hypernym-hyponym relationships, the repository contains over 300M models, with 220,000 classified into 3,135 classes. There are 31,693 meshes in the ShapeNet Parts subset, divided into 16 categories of objects (\textit{i.e.}, tables, chairs, planes, \textit{etc.}). Each shape contains 2-5 parts (with 50 part classes in total).

\textbf{GLUE}~\citep{wang2018glue}:
General Language Understanding Evaluation (GLUE) benchmark is a collection of nine natural language understanding tasks, including single-sentence tasks CoLA and SST-2, similarity and paraphrasing tasks MRPC, STS-B and QQP, and natural language inference tasks MNLI, QNLI, RTE, and WNLI.
Among them, SST-2, MRPC, QQP, MNLI, QNLI, RTE, and WNLI use accuracy as the metric, which is defined in CIFAR10. CoLA is measured by Matthews Correlation Coefficient (MCC), which is better in binary classification since the number of positive and negative samples are extremely unbalanced:

\begin{equation}
    MCC = \frac{TP \times TN - FP \times FN}{\sqrt{(TP+FP)(TP+FN)(TN+FP)(TN+FN)}}.
\end{equation}

And STS-B is measured by Pearson/Spearman Correlation Coefficient:

\begin{equation}
    r_{\textit{Pearson}} = \frac{1}{n-1} \sum^n_{i=1} \left( \frac{X_i - \bar{X}}{s_X} \right)  \left( \frac{Y_i - \bar{Y}}{s_Y} \right),
    r_{\textit{Spearman}} = 1-\frac{6\sum d^2_i}{n(n^2-1)},
\end{equation}

where $n$ is the number of observations, $s_X$ and $s_Y$ indicate the sum of squares of $X$ and $Y$ respectively, and $d_i$ is the difference  between the ranks of corresponding variables. 

\textbf{SpeechCom.}~\citep{warden2018speech}:
As part of its training and evaluation process, SpeechCom provides a collection of audio recordings containing spoken words. Its primary goal is to provide a way to build and test small models that detect a single word that belongs to a set of ten target words. Models should detect as few false positives as possible from background noise or unrelated speech while providing as few false positives as possible.
The accuracy metric for SpeechCom is also the same as CIFAR10. 


\textbf{CIFAR10-C}~\citep{hendrycks2018benchmarking}:
CIFAR10-C is a dataset generated by adding 15 common corruptions and 4 extra corruptions to the test images in the Cifar10 dataset. It benchmarks the frailty of classifiers under corruption, including noise, blur, weather, and digital influence. And each type of corruption has five levels of severity, resulting in 75 distinct corruptions. 
We report the accuracy of the classifiers under each level of severity and each corruption. Meanwhile, we use the mean and relative corruption error as metrics. Denote the error rate of $\textit{Network}$ under $\textit{Settings}$ as $E^{\textit{Network}}_{\textit{Settings}}$. The classifier's aggregate performance across the five severities of the corruption types. The Corruption Errors of a certain type of $\textit{Corruption}$ is computed with the formula:

\begin{equation}
    CE^{\textit{Network}}_{\textit{Corruption}} = \sum^5_{s=1} E^{\textit{Network}}_{s, \textit{Corruption}} / \sum^5_{s=1} E^{\textit{AlexNet}}_{s, \textit{Corruption}}.
\end{equation}

To make Corruption Errors comparable across corruption types, the difficulty is usually adjusted by dividing by AlexNet's errors.

\subsection{Details of Neural Architectures}
\label{app:arch}

\textbf{ResNet}~\citep{he2016deep}:
Residual Networks, or ResNets, learn residual functions concerning the layer inputs instead of learning unreferenced functions. Instead of making stacked layers directly fit a desired underlying mapping, residual nets let these layers fit a residual mapping. There is empirical evidence that these networks are easier to optimize and can achieve higher accuracy with considerably increased depth. 

\textbf{VGG}~\citep{simonyan2015very}:
VGG is a classical convolutional neural network architecture. 
It is proposed by an analysis of how to increase the depth of such networks. It is characterized by its simplicity: the network utilizes small 3$\times$3 filters, and the only other components are pooling layers and a fully connected layer.

\textbf{MobileNetV2}~\citep{mobilenet_v2}:
MobileNetV2 is a convolutional neural network architecture that performs well on mobile devices. This model has an inverted residual structure with residual connections between the bottleneck layers. The intermediate expansion layer employs lightweight depthwise convolutions to filter features as a source of nonlinearity. In MobileNetV2, the architecture begins with an initial layer of 32 convolution filters, followed by 19 residual bottleneck layers. 

\textbf{Faster-RCNN}~\citep{NIPS2015_5638}:
Faster R-CNN is an object detection model that improves Fast R-CNN by utilizing a region proposal network (RPN) with the CNN model. The RPN shares full-image convolutional features with the detection network, enabling nearly cost-free region proposals. A fully convolutional network is used to predict the bounds and objectness scores of objects at each position simultaneously. RPNs use end-to-end training to produce region proposals of high quality and instruct the unified network where to search. Sharing their convolutional features allows RPN and Fast R-CNN to be combined into a single network. Faster R-CNN consists of two modules. The first module is a deep, fully convolutional network that proposes regions, and the second is the detector that uses the proposals for giving the final prediction boxes. 

\textbf{SSD}~\citep{liu2016ssd}:
SSD is a single-stage object detection method that discretizes the output space of bounding boxes into a set of default boxes over different aspect ratios and scales per feature map location. During prediction, each default box is adjusted to match better the shape of the object based on its scores for each object category. In addition, the network automatically handles objects of different sizes by combining predictions from multiple feature maps with different resolutions.

\textbf{BERT}~\citep{kenton2019bert}:
BERT, or Bidirectional Encoder Representations from Transformers, improves upon standard Transformers by removing the unidirectionality constraint using a masked language model (MLM) pre-training objective. 
By masking some tokens from the input, the masked language model attempts to estimate the original vocabulary id of the masked word based solely on its context. An MLM objective differs from a left-to-right language model in that it enables the representation to integrate the left and right contexts, which facilitates pre-training a deep bidirectional Transformer. Additionally, BERT uses a next-sentence prediction task that pre-trains text-pair representations along with the masked language model. Note that we replace the direct binarized attention with a bi-attention mechanism to prevent the model from completely crashing~\citep{qin2021bibert}.

\textbf{PointNet}~\citep{qi2017pointnet}:
PointNet is a unified architecture for applications ranging from object classification and part segmentation to scene semantic parsing. The architecture directly receives point clouds as input and outputs either class labels for the entire input or point segment/part labels. \textbf{PointNet-Vanilla} is a variant of PointNet, which drops off the T-Net module. And for all PointNet models, we apply the EMA-Max~\citep{qin2020bipointnet} as the aggregator, because directly following the max pooling aggregator will cause the binarized PointNets to fail to converge.


\textbf{FSMN}~\citep{zhang2015feedforward}:
Feedforward sequential memory networks or FSMN is a novel neural network structure to model long-term dependency in time series without using recurrent feedback. It is a standard fully connected feedforward neural network containing some learnable memory blocks. As a short-term memory mechanism, the memory blocks encode long context information using a tapped-delay line structure.

\textbf{Deep-FSMN}~\citep{zhang2018deep}:
The Deep-FSMN architecture is an improved feedforward sequential memory network (FSMN) with skip connections between memory blocks in adjacent layers. By utilizing skip connections, information can be transferred across layers, and thus the gradient vanishing problem can be avoided when building very deep structures.

\subsection{Details of Hardware}
\label{app:hardware}

\textbf{Hisilicon Kirin}~\citep{kirin}:
Kirin is a series of ARM-based systems-on-a-chip (SoCs) produced by HiSilicon. Their products include Kirin 970, Kirin 980, Kirin 985, \textit{etc.}

\textbf{MediaTek Dimensity}~\citep{mediatek}:
Dimensity is a series of ARM-based systems-on-a-chip (SoCs) produced by MediaTek. Their products include Dimensity 820, Dimensity 8100, Dimensity 9000, \textit{etc.}

\textbf{Qualcomm Snapdragon}~\citep{singh2014evolution}:
Snapdragon is a family of mobile systems-on-a-chip (SoC) processor architecture provided by Qualcomm. 
The original Snapdragon chip, the Scorpio, was similar to the ARM Cortex-A8 core based upon the ARMv7 instruction set, but it was enhanced by the use of SIMD operations, which provided higher performance. Qualcomm Snapdragon processors are based on the Krait architecture. They are equipped with an integrated LTE modem, providing seamless connectivity across 2G and 3G LTE networks. 

\textbf{Raspberrypi}~\citep{raspberrypi}:
Raspberry Pi is a series of small single-board computers (SBCs) developed in the United Kingdom by the Raspberry Pi Foundation in association with Broadcom. 
Raspberry Pi was originally designed to promote the teaching of basic computer science in schools and in developing countries. As a result of its low cost, modularity, and open design, it is used in many applications, including weather monitoring, and is sold outside the intended market. It is typically used by computer hobbyists and electronic enthusiasts due to the adoption of HDMI and USB standards.

\textbf{Apple M1}~\citep{applem1}:
Apple M1 is a series of ARM-based systems-on-a-chip (SoCs) designed by Apple Inc. 
As a central processing unit (CPU) and graphics processing unit (GPU) for Macintosh desktops and notebooks, as well as iPad Pro and iPad Air tablets. In November 2020, Apple introduced the M1 chip, followed by the professional-oriented M1 Pro and M1 Max chips in 2021. Apple launched the M1 Ultra in 2022, which combines two M1 Max chips in a single package. The M1 Max is a higher-performance version of the M1 Pro, with larger GPU cores and memory bandwidth.

\subsection{Discussion of Novelty and Significance}
\label{subsec:DiscussionofNovelty}
We emphasize that our BiBench includes the following significant contributions: (1) the \textit{\textbf{first}} systematic benchmark that enables a new view to quantitative evaluate binarization algorithms at the operator level, and (2) revealing a practical binarized operator design paradigm.

(1) BiBench is the first effort to facilitate systematic and comprehensive comparisons between binarized algorithms. It provides a brand new perspective to decouple the binarized operators from the architectures for quantitative evaluations at the operator level. In existing works, the binarized operator and specific structure/training pipeline are often designed simultaneously or coupled closely (up to 15 (in 26) algorithms in Table 5 of the manuscript), the latter is often difficult to generalize to various architectures and tasks. Their training and inference settings are also different from each other. This makes the direct comparison among binarization algorithms difficult. Our BiBench enables a new approach towards a fair comparison of binarization algorithms by building a unified evaluation track for each algorithm on learning tasks and neural architectures. 

(2) BiBench reveals a practical paradigm of binarized operator designing. Based on the systemic and quantitative evaluation, superior techniques for better binarization operators can emerge, which is essential for pushing binarization algorithms to be accurate and efficient. For instance, after excluding the bi-real shortcut in Bi-Real Net, the difference between it and the DoReFa operator is solely the soft estimator in the backward propagation, yet this change yields a 2\% difference in the learning task track (as Table~\ref{tab:acc1} in the manuscript shows). These unprecedented quantitative insights identify which techniques are effective and low-cost for improving binarization operators, which will strongly facilitate the emergence of more powerful generic binarization algorithms. We summarize and present these operator-level binarization techniques in Section~\ref{subsec:SuggestedParadigm}.

\subsection{Discussion of Bit-width Constraints}
In real practice, applying flexible bit-width or specialized quantizers in certain layers can lead to significant improvements in binarization, enabling a better balance between task performance and inference efficiency. There is a common practice to keep the input and output layers at 32-bit full precision since binarizing these two layers brings a severe task performance drop while having a relatively low efficiency gain~\cite{rastegari2016xnor,Bi-Real,Dorefa-Net}. And we also follow this practice in our BiBench. Moreover, some binarization algorithms optimized for practical deployment also compress these layers while maintaining the task performance, they usually make the two layers fixed-point (\textit{e.g.}, 8-bit) or design specific quantizers for them~\cite{zhang2021fracbnn,zhao2017accelerating,courbariaux2016binarized}. As mentioned by the reviewer, FracBNN~\cite{zhang2021fracbnn} is one of the most representative works among them, which employs a dual-precision activation scheme to compute features with up to two bits using an additional sparse binary convolution. In practice, these algorithms can help BNNs achieve better accuracy-efficiency trade-offs.

We would also highlight that there is significant value in the exploration of full binarization (weight and activation) algorithms at the operator level, which is the focus of our BiBench. Since the application of the most aggressive bit-widths, the binarized parts in BNNs can be implemented by extremely efficient bitwise instructions, which makes it a more efficient solution compared to other bit-width quantization and partial binarization. Hence, the weight and activation binarization is adopted by various papers that propose binarization algorithms as a standard setting~\cite{rastegari2016xnor,Dorefa-Net,Bi-Real,liu2020reactnet}. Therefore, our BiBench applies binarization operators to the entire network to allow us to fairly reflect the accuracy and efficiency of various binarization operators. Furthermore, the goal of BiBench is to systematically study various binarization algorithms through the results of each evaluation track (including learning tasks, neural architecture, \textit{etc}.). So we only consider generic binarization techniques at the operator level when benchmarking, while techniques like special structural design and quantization for specific layers are excluded. We believe that our results provide valuable insights into the practical trade-offs and challenges of using binary activations, which can serve as a foundation for future research in this area.

\begin{table}[t]
    \centering
    \caption{Results of the structure binarization.}
    \label{tab:discussion_bi_arch}
    \vspace{-0.1in}
    \setlength{\tabcolsep}{5.5mm}{
    \begin{tabular}{lcccccc}
    \toprule
        ~ & Size$\backslash$Dim & 64 & 128 & 256 & 512 & Mean \\ 
        \midrule
        Attention (Token) & 64 & 0.97 & 0.99 & 1.02 & 1.03 & 0.98 \\ 
        ~ & 128 & 0.97 & 0.98 & 1.01 & 1.03 & ~ \\ 
        ~ & 256 & 0.94 & 0.96 & 0.99 & 1.01 & ~ \\ 
        ~ & 512 & 0.89 & 0.92 & 0.97 & 0.99 & ~ \\ 
        \midrule
        CNN (W$\times$H) & 14$\times$14 & 0.59 & 0.59 & 0.59 & 0.59 & 0.59 \\ 
        ~ & 28$\times$28 & 0.59 & 0.58 & 0.59 & 0.59 & ~ \\ 
        ~ & 56$\times$56 & 0.59 & 0.59 & 0.59 & 0.59 & ~ \\ 
        ~ & 112$\times$112 & 0.59 & 0.59 & 0.59 & 0.59 & ~ \\ 
        \midrule
        MLP (\#Point) & 64 & 0.84 & 0.85 & 0.84 & 0.86 & 0.84 \\ 
        ~ & 128 & 0.84 & 0.84 & 0.85 & 0.84 & ~ \\ 
        ~ & 256 & 0.84 & 0.84 & 0.84 & 0.84 & ~ \\
        ~ & 512 & 0.84 & 0.84 & 0.84 & 0.84 & ~ \\ 
        \bottomrule
    \end{tabular}}
\end{table}

\subsection{Discussion of Binarized Architectures}
\label{subsec:DiscussionofBinarizedArchitectures}
To empirically verify our analysis in Section~\ref{subsubsec:NeuralArchitectureBinarizationonTransformers}, we design the following experiments to demonstrate that the impact of different architectures on binarization is mainly caused by different local structures: (1) Local structures definition. We first abstract the typical local structures of CNN-based, transformer-based, and MLP-based architectures. The first is the bottleneck unit in ResNet, which is mainly composed of three 2D convolutions and a shortcut. The second is the self-attention structure in BERT, which is mainly composed of three linear units generating the query, key, and value, and two multiplication operations between activations (without weights). The last one is the sub-MLP structure in PointNet, which is mainly formed by stacking 3 linear units in series. All binarized structure examples unified contain three convolution or linear units to ensure consistency. (2) Initialization and binarization. To get the impact of the structure level on binarization, we randomly initialize the weights and inputs in all structures and then compare the differences between their full precision and binarized versions. We use the most basic BNN binarization to reveal essential effects. (3) Metric definition. We define 16 inputs of different sizes for each structure, and compare the average error caused by binarization on them:

\begin{equation}
E_f=\mathbb{E}_{\boldsymbol{x}\in\mathcal{X}}\left(\left|\frac{f(\boldsymbol{x})}{\operatorname{std}(f(\boldsymbol{x}))}-\frac{\hat{f}(\boldsymbol{x})}{\operatorname{std}(\hat{f}(\boldsymbol{x}))}\right| _{\ell1}\right),
\end{equation}

where $\boldsymbol{x}\in\mathcal{X}$ denotes the input set, $\|\cdot\| _{\ell1}$ denotes L1 norm, and $f(\cdot)$ and $\hat{f}(\cdot)$ denote the full-precision and binarized structures, respectively. As shown in the results in Table~\ref{tab:discussion_bi_arch}, the transformer exhibits a larger binarization error at the local structure level. This empirically validates our analysis that binarization has a greater impact on the forward propagation of transformers with the attention mechanism compared to other architectures.

\subsection{Discussion of Binarization Robustness}
\label{subsec:DiscussionofBinarization}
we first compare corruption robustness between full-precision architectures of different sizes to determine whether smaller model sizes necessarily lead to better robustness. We compare the full-precision ResNet-20 evaluated by BiBench. ResNet-20 has a consistent residual structure compared with ResNet-18, and the latter has a smaller parameter amount (about 1/40 of the former). However, our results in Table~\ref{tab:robustness-indicators} counter-intuitively show that the robustness indicator of full-precision ResNet-20 is only 89.55 in terms of robustness, which is worse than not only up to 7 (out of 8) binarization methods we evaluated but also the larger-scale full-precision ResNet-18 (whose corresponding indicator is 100.00). This suggests that the reason for being robust to data corruption cannot be directly attributed to the smaller model scale, but is more likely to be a unique characteristic of some binarization algorithms. More detailed results are in Table~\ref{tab:originalresultsof}.

\begin{table}[t]
    \centering
    \caption{Robustness indicators of binarization algorithms and full-precision ResNet-20.}
    \label{tab:robustness-indicators}
    \vspace{-0.1in}
    \setlength{\tabcolsep}{1.5mm}
    {\begin{tabular}{lccccccccc}
    \toprule
        Arch & BNN & XNOR & DoReFa & Bi-Real & XNOR++ & ReActNet & ReCU & FDA & ResNet 20 (FP32) \\ \midrule
        CIFAR10-C & 95.26 & 100.97 & 81.43 & 96.56 & 92.69 & 94.01 & 103.29 & 98.35 & 89.55 \\ 
    \bottomrule
    \end{tabular}}
\end{table}

\begin{table}[!ht]
    \centering
    \caption{Original results of full-precision ResNet-20 on CIFAR10-C.}
    \label{tab:originalresultsof}
    \vspace{-0.1in}
    \resizebox{\linewidth}{!}{
    \begin{tabular}{llllllllllllllllllll}
    \toprule
        Origin & 91.99 & Overall & 68.68 & ~ & ~ & ~ & ~ & ~ & ~ & ~ & ~ & ~ & ~ & ~ & ~ & ~ & ~ & ~ & ~ \\ \midrule
        gaussian\_noise1 & 69.91 & shot\_noise1 & 80.41 & gaussian\_blur1 & 91.44 & glass\_blur1 & 47.34 & zoo\_blur1 & 81.48 & fog1 & 91.71 & snow1 & 84.26 & elastic\_transfor1 & 85.65 & pixelate1 & 88.95 & spatter1 & 88.02 \\ 
        gaussian\_noise2 & 48.26 & shot\_noise2 & 68.26 & gaussian\_blur2 & 81.83 & glass\_blur2 & 48.49 & zoo\_blur2 & 76.91 & fog2 & 89.92 & snow2 & 70.81 & elastic\_transfor2 & 85.39 & pixelate2 & 82.4 & spatter2 & 80.3 \\ 
        gaussian\_noise3 & 32.12 & shot\_noise3 & 44.64 & gaussian\_blur3 & 67.53 & glass\_blur3 & 51.63 & zoo\_blur3 & 69.56 & fog3 & 87.18 & snow3 & 75.67 & elastic\_transfor3 & 80.06 & pixelate3 & 76.87 & spatter3 & 76.17 \\ 
        gaussian\_noise4 & 27.33 & shot\_noise4 & 37.75 & gaussian\_blur4 & 52.48 & glass\_blur4 & 40.26 & zoo\_blur4 & 62.59 & fog4 & 82.64 & snow4 & 72.39 & elastic\_transfor4 & 73.32 & pixelate4 & 59.62 & spatter4 & 79.82 \\ 
        gaussian\_noise5 & 23.29 & shot\_noise5 & 28.72 & gaussian\_blur5 & 31.23 & glass\_blur5 & 41.77 & zoo\_blur5 & 52.02 & fog5 & 61.65 & snow5 & 64.81 & elastic\_transfor5 & 69.55 & pixelate5 & 43.48 & spatter5 & 69.63 \\ 
        ipulse\_nosie1 & 80.6 & speckle\_noise1 & 80.58 & defocus\_blur1 & 91.39 & otion\_blur1 & 85.17 & brighness1 & 91.78 & frost1 & 85.77 & contrast1 & 91.37 & jpeg\_copression1 & 83.15 & saturate1 & 89.24 & ~ & ~ \\ 
        ipulse\_nosie2 & 67.92 & speckle\_noise2 & 63.85 & defocus\_blur2 & 89.24 & otion\_blur2 & 75.86 & brighness2 & 91.27 & frost2 & 78.21 & contrast2 & 85.99 & jpeg\_copression2 & 77.7 & saturate2 & 86.71 & ~ & ~ \\ 
        ipulse\_nosie3 & 58.22 & speckle\_noise3 & 56.01 & defocus\_blur3 & 81.91 & otion\_blur3 & 65.86 & brighness3 & 90.52 & frost3 & 65.87 & contrast3 & 79.12 & jpeg\_copression3 & 75.65 & saturate3 & 91.01 & ~ & ~ \\ 
        ipulse\_nosie4 & 43.37 & speckle\_noise4 & 44.78 & defocus\_blur4 & 70.09 & otion\_blur4 & 65.56 & brighness4 & 89.4 & frost4 & 63.54 & contrast4 & 64.85 & jpeg\_copression4 & 72.6 & saturate4 & 87.96 & ~ & ~ \\ 
        ipulse\_nosie5 & 31.27 & speckle\_noise5 & 34.8 & defocus\_blur5 & 47.18 & otion\_blur5 & 56.16 & brighness5 & 86.63 & frost5 & 51.87 & contrast5 & 22.13 & jpeg\_copression5 & 69.08 & saturate5 & 83.69 & ~ & ~ \\ \bottomrule
    \end{tabular}}
\end{table}

\section{Full Results}
\label{sec:FullResults}

\subsection{Evaluation Results of All Tracks}
Table~\ref{tab:appendix-vision-acc}-\ref{tab:appendix-shapenet-acc} shows the accuracy of different binarization algorithms on 2D and 3D vision tasks, including CIFAR10, ImageNet, PASCAL VOC07, COCO17 for 2D vision tasks and ModelNet40 for 3D vision task. And for each task, we cover several representative model architectures and binarize them with the binarization algorithms mentioned above. 

We also evaluate binarization algorithms on language and speech tasks, for which we test TinyBERT (4 layers and 6 layers) on GLUE Benchmark and FSMN and D-FSMN on the SpeechCommand dataset. Results are listed in Table~\ref{tab:appendix-language-acc}. 

To demonstrate the robustness corruption of binarized algorithms, we show the results on the CIFAR10-C benchmark, which is used to benchmark the robustness to common perturbations in Table~\ref{tab:appendix-corr-acc} and Table~\ref{tab:appendix-corr-acc2}. It includes 15 kinds of noise, blur, weather, and digital corruption, each with five levels of severity. 

The sensitivity of hyperparameters while training is shown in Table~\ref{tab:appendix-sensitivity-acc1}-\ref{tab:appendix-sensitivity-acc2}. For each binarization algorithm, we use SGD or Adam optimizer, 1$\times$ or 0.1$\times$ of the original learning rate, cosine or step learning scheduler, and 200 training epochs. Each case is tested five times to show the training stability. We also calculate the mean and standard deviation (std) of accuracy. The best accuracy and the lowest std for each binarization algorithm are bolded. 

We conduct comprehensive deployment and inference on various kinds of hardware, including the Kirin series (970, 980, 985, 990, and 9000E), Dimensity series (820 and 9000), Snapdragon series (855+, 870 and 888), Raspberrypi (3B+ and 4B) and Apple M1 series (M1 and M1 Max). Limited to the support of frameworks, we can only test BNN and ReAct with Larq compute engine and only BNN with daBNN. We convert models to enable the actual inference on real hardware, including ResNet18/34 and VGG-Small on Larq, and only ResNet18/34 on daBNN. And we test 1, 2, 4, and 8 threads for each hardware and additionally test 16 threads for Apple Silicons on Larq. And daBNN only supports single-thread inference. Results are showcased in  Table~\ref{app:tab:eff1}-\ref{app:tab:eff4}.

\begin{table}[t]
\renewcommand\arraystretch{2.7}
\centering
\caption{Accuracy on 2D and 3D Vision Tasks.}
\label{tab:appendix-vision-acc}
\vspace{-0.1in}
\setlength{\tabcolsep}{1.5mm}
{
\begin{tabular}{lllccccccccccc}
\toprule
\multirow{2}{*}{Task} & & \multirow{2}{*}{Arch.} & \multirow{2}{*}{FP32} & \multicolumn{8}{c}{Binarization Algorithm} \\
 & & & & {BNN} & {XNOR} & {DoReFa} & {Bi-Real} & {XNOR++} & {ReActNet} & {ReCU} & {FDA} \\
\midrule
\multirow{4}{*}{CIFAR10} & & ResNet20 & 91.99 & 85.31 & 85.53 & 85.18 & 85.56 & 85.41 & 86.18 & 86.42 & 86.38  \\
 & & ResNet18 & 94.82 & 89.69 & 91.4 & 91.55 & 91.20 & 90.04 & 91.55 & 92.79 & 90.42 \\
 & & ResNet34 & 95.34 & 90.82 & 89.58 & 90.95 & 92.50 & 90.59 & 92.69 & 93.64 & 89.59  \\
 & & VGG-Small & 93.80 & 89.66 & 89.65 & 89.66 & 90.25 & 89.34 & 90.27 & 90.84 & 89.48  \\
\midrule
{ImageNet} & & ResNet18 & 69.90 & 52.99 & 53.99 & 53.55 & 54.79 & 52.43 & 54.97 & 54.51 & 54.63 	\\
\midrule
\multirow{2}{*}{VOC07} & & Faster-RCNN  & 76.06 & 58.54 & 56.75 & 58.07 & 60.90 & 56.60 & 61.90 & 62.10 & 60.10 \\
& & SSD300 & 77.34 & 9.09 & 33.72 & 30.70 & 31.90 & 9.41 & 38.41 & 9.80 & 43.68 \\
\midrule
{COCO17} & & Faster-RCNN & 27.20 & 21.20 & 20.50 & 21.30 & 22.20 & 21.60 & 22.80 & 23.30 & 22.40 \\
\midrule
\multirow{2}{*}{ModelNet40} & &  PointNet$_\textrm{vanilla}$ & 86.80 & 85.13 & 83.47 & 85.21 & 85.37 & 85.66 & 85.13 & 85.21 & 85.49\\
 & & PointNet & 88.20 & 9.08 & 80.75 & 78.77 & 77.71 & 63.25 & 76.50 & 81.12 & 79.62 \\
\bottomrule
\end{tabular}}
\end{table}

\begin{table}[t]
\renewcommand\arraystretch{2.7}
\centering
\caption{Accuracy on ShapeNet dataset.}
\label{tab:appendix-shapenet-acc}
\vspace{-0.1in}
\setlength{\tabcolsep}{1.mm}
{
\begin{tabular}{llllccccccccccc}
\toprule
\multirow{2}{*}{Task} & \multirow{2}{*}{Arch.} & \multirow{2}{*}{Category} & \multirow{2}{*}{FP32} & \multicolumn{8}{c}{Binarization Algorithm} \\
 & & & & {BNN} & {XNOR} & {DoReFa} & {Bi-Real} & {XNOR++} & {ReActNet} & {ReCU} & {FDA} \\
\midrule
\multirow{17}{*}{ShapeNet} & \multirow{17}{*}{PointNet} & Airplane & 83.7 & 37.5 & 74.14 & 67.2 & 67.61 & 30.36 & 66.12 & 31.61 & 65.34 \\ 
        ~ & ~ & Bag & 79.6 & 44.2 & 49 & 55.34 & 47.11 & 37.44 & 50.28 & 38.58 & 48.62 \\ 
        ~ & ~ & Cap & 92.3 & 44.3 & 73.32 & 51.21 & 61.41 & 40.37 & 56.73 & 40.13 & 56 \\ 
        ~ & ~ & Car & 76.8 & 24.3 & 55.27 & 52.24 & 49.39 & 24.07 & 49.11 & 23.92 & 58.5 \\ 
        ~ & ~ & Chair & 90.9 & 61.6 & 85.62 & 83.96 & 83.6 & 41.89 & 83.83 & 41.5 & 83.27 \\ 
        ~ & ~ & Earphone & 70.2 & 38.5 & 30.97 & 34.94 & 35.24 & 26.3 & 36.72 & 23.01 & 34.46 \\ 
        ~ & ~ & Guitar & 91.1 & 32.9 & 69.17 & 67.9 & 65.99 & 23.45 & 64.18 & 28.38 & 78.69 \\ 
        ~ & ~ & Knife & 85.7 & 43 & 78.16 & 76.16 & 75.53 & 37.62 & 75.01 & 38.81 & 77.07 \\ 
        ~ & ~ & Lamp & 82 & 51.2 & 69 & 68.75 & 60 & 49.45 & 66.13 & 48.41 & 67.45 \\ 
        ~ & ~ & Laptop & 95.5 & 49.4 & 93.29 & 92.93 & 92.79 & 41.89 & 92.93 & 42.28 & 93.66 \\ 
        ~ & ~ & Motorbike & 64.4 & 16.3 & 19.04 & 18.88 & 18.69 & 13.18 & 18.59 & 11.26 & 20.38 \\ 
        ~ & ~ & Mug & 93.6 & 49.1 & 64.32 & 53.56 & 52.01 & 47.58 & 52.51 & 46.83 & 53.48 \\ 
        ~ & ~ & Pistol & 80.8 & 25.5 & 62.29 & 59.15 & 51.43 & 26.96 & 53.79 & 27.81 & 62.61 \\ 
        ~ & ~ & Rocket & 54.4 & 26.9 & 30.95 & 27.92 & 26.61 & 22.38 & 26.01 & 19.32 & 23.08 \\ 
        ~ & ~ & Skateboard & 70.7 & 41.2 & 45.7 & 50.15 & 43.78 & 28.63 & 43.74 & 26.71 & 45.81 \\ 
        ~ & ~ & Table & 81.4 & 51.3 & 73.68 & 72.69 & 69.72 & 45.74 & 69.56 & 45.21 & 73.45 \\ 
        ~ & ~ & Overall & 80.81875 & 39.82 & 60.87 & 58.31 & 56.31 & 33.58 & 56.58 & 33.36 & 58.68 \\ 
\bottomrule
\end{tabular}}
\end{table}

\begin{table}[h]
\renewcommand\arraystretch{1.5}
\centering
\caption{Accuracy on Language and Speech Tasks.}
\label{tab:appendix-language-acc}
\vspace{-0.1in}
\setlength{\tabcolsep}{1.2mm}
{
\begin{tabular}{lllcccccccccccc}
\toprule
\multicolumn{2}{l}{\multirow{2}{*}{Task}} & \multirow{2}{*}{Arch.} & \multirow{2}{*}{FP32} & \multicolumn{8}{c}{Binarization Algorithm} \\
& & & & {BNN} & {XNOR} & {DoReFa} & {Bi-Real} & {XNOR++} & {ReActNet} & {ReCU} & {FDA} \\
\midrule
\multirow{27}{*}{GLUE} & \multirow{3}{*}{MNLI-m} & BERT-Tiny$_\textrm{4L}$ & 82.81 & 36.90 & 41.20 & 52.31 & 55.09 & 37.27 & 55.52 & 38.55 & 59.41 \\
 &  & BERT-Tiny$_\textrm{6L}$ & 84.76 & 37.01 & 51.17 & 63.09 & 66.81 & 37.98 & 66.47 & 37.95 & 68.46 \\
 &  & BERT-Base & 84.88 & 35.45 & 41.40 & 60.67 & 62.47 & 35.45 & 60.22 & 35.45 & 63.49 \\
\cmidrule{2-15}
 & \multirow{3}{*}{MNLI-mm}  & BERT-Tiny$_\textrm{4L}$ & 83.08 & 36.54 & 41.55 & 53.01 & 55.57 & 36.07 & 55.89 & 37.62 & 59.76 \\
 &  & BERT-Tiny$_\textrm{6L}$ & 84.42 & 36.47 & 50.92 & 63.87 & 66.82 & 38.11 & 67.64 & 36.91 & 68.98 \\
 &  & BERT-Base & 85.45 & 35.22 & 41.18 & 60.96 & 63.17 & 35.22 & 61.19 & 35.22 & 63.72 \\
\cmidrule{2-15}
 &  \multirow{3}{*}{QQP}  & BERT-Tiny$_\textrm{4L}$ & 90.47 & 66.19 & 73.69 & 75.79 & 77.38 & 64.97 & 76.92 & 67.32 & 78.92\\
 &  & BERT-Tiny$_\textrm{6L}$ & 85.98 & 63.18 & 78.90 & 80.93 & 82.42 & 63.19 & 82.95 & 63.3 & 83.19\\
 &  & BERT-Base & 91.51 & 63.18 & 71.93 & 77.07 & 80.01 & 63.18 & 81.16 & 63.18 & 83.26 \\
\cmidrule{2-15}
 & \multirow{3}{*}{QNLI}  & BERT-Tiny$_\textrm{4L}$ & 87.46 & 51.71 & 60.59 & 61.15 & 61.92 & 52.79 & 62.67 & 53.99 & 62.29 \\
 &  & BERT-Tiny$_\textrm{6L}$ & 90.79 & 52.22 & 62.75 & 66.88 & 69.72 & 51.84 & 70.27 & 51.32 & 72.72 \\
 &  & BERT-Base & 92.14 & 51.8 & 60.29 & 70.78 & 70.14 & 54.07 & 69.44 & 51.87 & 72.43 \\
\cmidrule{2-15}
 & \multirow{3}{*}{SST-2}  & BERT-Tiny$_\textrm{4L}$ & 92.43 & 52.98 & 79.93 & 82.45 & 84.06 & 54.01 & 84.17 & 54.24 & 86.12\\
 &  & BERT-Tiny$_\textrm{6L}$ & 90.25 & 58.14 & 84.74 & 86.23 & 87.73 & 69.38 & 87.95 & 52.40 & 87.72 \\
 &  & BERT-Base & 93.23 & 52.29 & 78.78 & 86.01 & 86.35 & 53.32 & 84.4 & 52.40 & 87.93 \\
\cmidrule{2-15}
 & \multirow{3}{*}{CoLA}  & BERT-Tiny$_\textrm{4L}$ & 49.61 & 6.55 & 7.22 & 12.69 & 16.86 & 0 & 14.71 & 6.25 & 17.80  \\
 &  & BERT-Tiny$_\textrm{6L}$ & 54.17 & 2.57 & 12.57 & 15.97 & 17.94 & 0 & 15.24 & 2.24 & 22.21  \\
 &  & BERT-Base & 59.71 & 4.63 & 0 & 4.74 & 15.95 & 0 & 4.63 & 0.40 & 4.63 \\
\cmidrule{2-15}
 & \multirow{3}{*}{STS-B}  & BERT-Tiny$_\textrm{4L}$ & 86.35 & 4.31 & 18.05 & 18.74 & 22.65 & 7.45 & 22.73 & 8.20 & 27.56 \\
 &  & BERT-Tiny$_\textrm{6L}$ & 89.79 & 1.04 & 14.72 & 22.31 & 24.59 & 5.70 & 23.40 & 8.22 & 37.15 \\
 &  & BERT-Base & 90.06 & 6.94 & 12.19 & 18.26 & 20.76 & 4.99 & 8.73 & 6.59 & 10.14  \\
\cmidrule{2-15}
 & \multirow{3}{*}{MRPC}  & BERT-Tiny$_\textrm{4L}$ & 85.50 & 68.30 & 71.74 & 71.99 & 71.74 & 68.30 & 71.74 & 71.25 & 71.49 \\
 &  & BERT-Tiny$_\textrm{6L}$ & 87.71 & 68.30 & 70.76 & 71.74 & 71.49 & 68.30 & 71.74 & 69.04 & 71.74  \\
 &  & BERT-Base & 86.24 & 68.30 & 68.3 & 70.02 & 70.27 & 68.30 & 71.25 & 68.30 & 69.04 \\
\cmidrule{2-15}
 & \multirow{3}{*}{RTE}  & BERT-Tiny$_\textrm{4L}$ & 65.34 & 56.31 & 53.43 & 56.31 & 55.59 & 54.15 & 57.76 & 61.01 & 59.20 \\
 &  & BERT-Tiny$_\textrm{6L}$ & 68.95 & 56.31 & 54.51 & 54.51 & 58.12 & 49.09 & 53.43 & 58.84 & 54.87 \\
 &  & BERT-Base & 72.20 & 53.43 & 57.04 & 55.23 & 54.51 & 54.87 & 54.51 & 55.23 & 55.23 \\
\midrule
\multicolumn{2}{l}{\multirow{2}{*}{Speech Commands}} & FSMN & 94.89 & 56.45 & 56.45 & 68.65 & 73.60 & 75.04 & 73.80 & 56.45 & 56.45 \\
 &  & D-FSMN & 97.51 & 88.32 & 92.03 & 78.92 & 85.11 & 56.77 & 83.80 & 92.11 & 93.91\\
\bottomrule
\end{tabular}}
\end{table}

\begin{table}[h]
\renewcommand\arraystretch{0.98}
\centering
\caption{Results for Robustness Corruption on CIFAR10-C Dataset with Different Binarization Algorithms (1/2). }
\label{tab:appendix-corr-acc}
\vspace{-0.1in}
\setlength{\tabcolsep}{2.5mm}
{
\begin{tabular}{llrrrrrrrrrrrr}
\toprule
\multirow{2}{*}{Noise} & \multirow{2}{*}{FP32} & \multicolumn{8}{c}{Binarization Algorithm} \\
 & & {BNN} & {XNOR} & {DoReFa} & {Bi-Real} & {XNOR++} & {ReActNet} & {ReCU} & {FDA} \\
 \midrule
Origin & 94.82 & 89.69 & 91.40 & 91.55 & 91.20 & 90.04 & 91.55 & 92.79 & 90.42 \\ 
gaussian\_noise-1 & 78.23 & 74.22 & 76.00 & 74.97 & 74.95 & 75.07 & 75.15 & 78.25 & 77.36\\ 
gaussian\_noise-2 & 56.72 & 56.73 & 62.44 & 55.94 & 58.33 & 57.52 & 55.97 & 61.32 &60.48 \\ 
gaussian\_noise-3 & 36.93 & 42.69 & 47.58 & 39.56 & 43.47 & 40.79 & 37.99 & 43.32 &44.26 \\ 
gaussian\_noise-4 & 31.03 & 38.35 & 41.43 & 33.24 & 36.65 & 34.68 & 31.47 & 35.91 & 37.30 \\ 
gaussian\_noise-5 & 25.54 & 34.05 & 36.22 & 28.66 & 31.78 & 30.13 & 25.49 & 30.19 & 32.09 \\ 
ipulse\_nosie-1 & 82.54 & 84.57 & 86.94 & 84.68 & 87.30 & 85.72 & 86.89 & 88.73 & 85.8 \\ 
ipulse\_nosie-2 & 70.12 & 77.13 & 80.74 & 77.35 & 81.14 & 79.62 & 80.25 & 82.80 & 80.3 \\ 
ipulse\_nosie-3 & 59.88 & 70.58 & 75.01 & 69.20 & 74.59 & 71.82 & 72.16 & 76.05 & 72.6\\ 
ipulse\_nosie-4 & 40.59 & 54.42 & 59.39 & 49.48 & 56.66 & 52.61 & 49.79 & 58.44 & 56.45 \\ 
ipulse\_nosie-5 & 26.03 & 39.86 & 41.54 & 32.72 & 37.28 & 35.12 & 28.42 & 38.26 & 39.98\\ 
shot\_noise-1 & 85.75 & 81.51 & 81.31 & 81.84 & 81.58 & 80.66 & 81.88 & 83.98 & 82.42 \\ 
shot\_noise-2 & 76.61 & 72.04 & 74.02 & 72.21 & 72.81 & 73.03 & 72.32 & 76.70 & 75.1\\ 
shot\_noise-3 & 52.21 & 53.90 & 57.08 & 50.66 & 54.59 & 53.76 & 51.31 & 57.22 & 56.56 \\ 
shot\_noise-4 & 44.13 & 47.58 & 51.29 & 43.59 & 48.36 & 46.64 & 44.21 & 48.78 & 48.91 \\ 
shot\_noise-5 & 32.73 & 39.93 & 40.79 & 33.80 & 38.50 & 36.47 & 31.79 & 36.46 & 37.8\\ 
speckle\_noise-1 & 86.30 & 81.29 & 81.94 & 80.93 & 80.77 & 81.14 & 82.17 & 84.17 & 82.62\\ 
speckle\_noise-2 & 71.94 & 68.07 & 70.14 & 67.5 & 69.22 & 69.35 & 68.26 & 72.94 & 71.70\\ 
speckle\_noise-3 & 64.47 & 62.12 & 64.13 & 60.24 & 63.44 & 62.50 & 61.14 & 66.89 & 64.27\\ 
speckle\_noise-4 & 49.81 & 51.93 & 53.77 & 47.93 & 52.75 & 50.59 & 48.39 & 54.13 & 52.40 \\ 
speckle\_noise-5 & 38.70 & 44.25 & 43.60 & 38.65 & 43.16 & 42.09 & 37.78 & 42.13 & 42.57\\ 
gaussian\_blur-1 & 94.17 & 89.03 & 90.5 & 89.33 & 90.56 & 89.00 & 91.05 & 92.16 & 89.33\\ 
gaussian\_blur-2 & 87.04 & 78.3 & 81.98 & 78.81 & 80.42 & 77.75 & 81.20 & 84.80 & 78.93\\ 
gaussian\_blur-3 & 75.15 & 67.74 & 68.27 & 67.67 & 68.16 & 64.54 & 67.42 & 73.62 & 66.29\\ 
gaussian\_blur-4 & 59.5 & 55.17 & 53.63 & 55.74 & 54.08 & 52.44 & 52.72 & 60.32 & 53.37\\ 
gaussian\_blur-5 & 36.03 & 37.31 & 33.96 & 37.50 & 37.54 & 36.77 & 34.08 & 39.22 & 34.93\\ 
defocus\_blur-1 & 94.2 & 88.73 & 91.06 & 89.1 & 90.32 & 88.91 & 90.92 & 91.98 & 89.58\\ 
defocus\_blur-2 & 92.75 & 85.97 & 88.99 & 86.59 & 88.31 & 85.58 & 87.91 & 90.47 & 87.01\\ 
defocus\_blur-3 & 87.38 & 79.02 & 82.43 & 78.88 & 80.71 & 77.58 & 80.88 & 84.85 & 79.52\\ 
defocus\_blur-4 & 76.99 & 69.13 & 71.02 & 68.29 & 68.33 & 65.96 & 68.42 & 74.40 & 68.22\\ 
defocus\_blur-5 & 52.09 & 48.85 & 51.99 & 48.82 & 49.17 & 48.45 & 46.92 & 55.70 & 48.27 \\ 
glass\_blur-1 & 54.93 & 56.57 & 51.72 & 57.94 & 56.78 & 57.29 & 56.27 & 58.82 & 58.92 \\ 
glass\_blur-2 & 56.37 & 57.93 & 53.46 & 60.42 & 59.21 & 59.32 & 58.03 & 60.25 & 60.56 \\ 
glass\_blur-3 & 59.21 & 61.43 & 56.98 & 64.11 & 61.72 & 62.41 & 60.39 & 62.84 & 63.32 \\ 
glass\_blur-4 & 45.65 & 46.50 & 42.72 & 48.48 & 47.19 & 47.83 & 46.88 & 49.09 & 49.23 \\ 
glass\_blur-5 & 49.19 & 49.52 & 46.40 & 52.06 & 49.83 & 50.02 & 49.08 & 51.14 & 51.82 \\ 
otion\_blur-1 & 89.40 & 81.57 & 83.27 & 82.00 & 83.11 & 81.48 & 84.19 & 86.21 & 82.83 \\ 
otion\_blur-2 & 81.95 & 71.52 & 74.71 & 73.38 & 72.48 & 70.99 & 74.35 & 77.75 & 74.09 \\ 
otion\_blur-3 & 72.48 & 61.87 & 66.21 & 63.86 & 63.39 & 61.57 & 63.85 & 68.31 & 64.36 \\ 
otion\_blur-4 & 72.79 & 62.40 & 66.18 & 63.94 & 62.84 & 62.03 & 64.54 & 67.88 & 64.13 \\ 
otion\_blur-5 & 63.91 & 54.14 & 57.98 & 56.07 & 55.90 & 54.35 & 55.60 & 59.71 & 56.65 \\ 
zoo\_blur-1 & 87.36 & 78.25 & 81.31 & 78.56 & 79.45 & 77.20 & 80.22 & 83.69 & 78.55 \\ 
zoo\_blur-2 & 83.89 & 74.84 & 77.73 & 75.39 & 75.88 & 72.83 & 75.74 & 80.46 & 74.72 \\ 
zoo\_blur-3 & 77.73 & 69.00 & 70.98 & 68.81 & 69.03 & 66.56 & 68.34 & 74.33 & 68.21 \\ 
zoo\_blur-4 & 71.39 & 64.12 & 65.21 & 63.79 & 62.81 & 61.01 & 62.47 & 68.67 & 62.58 \\ 
zoo\_blur-5 & 60.60 & 55.15 & 55.83 & 55.4 & 54.38 & 52.66 & 52.24 & 59.51 & 53.94 \\ 
brighness-1 & 94.31 & 89.29 & 90.84 & 89.53 & 90.8 & 89.30 & 90.97 & 92.06 & 89.74 \\ 
brighness-2 & 94.03 & 88.25 & 90.42 & 88.71 & 89.66 & 88.50 & 90.64 & 91.64 & 88.77 \\ 
brighness-3 & 93.53 & 87.40 & 89.38 & 87.38 & 89.17 & 87.31 & 89.63 & 90.72 & 87.84 \\ 
brighness-4 & 92.74 & 85.45 & 88.27 & 86.12 & 87.78 & 85.44 & 88.16 & 89.58 & 86.39 \\ 
brighness-5 & 90.36 & 80.95 & 85.22 & 81.65 & 83.79 & 80.99 & 84.45 & 86.53 & 82.04 \\ 
\bottomrule
\end{tabular}}
\end{table}

\begin{table}[h]
\renewcommand\arraystretch{1.05}
\centering
\caption{Results for Robustness Corruption on CIFAR10-C Dataset with Different Binarization Algorithms (2/2).}
\label{tab:appendix-corr-acc2}
\vspace{-0.1in}
\setlength{\tabcolsep}{2.35mm}
{
\begin{tabular}{llrrrrrrrrrrrr}
\toprule
\multirow{2}{*}{Noise} & \multirow{2}{*}{FP32} & \multicolumn{8}{c}{Binarization Algorithm} \\
 & & {BNN} & {XNOR} & {DoReFa} & {Bi-Real} & {XNOR++} & {ReActNet} & {ReCU} & {FDA} \\
 \midrule
fog-1 & 94.04 & 88.17 & 90.89 & 88.84 & 89.91 & 88.51 & 90.84 & 92.08 & 89.43 \\ 
fog-2 & 93.03 & 84.58 & 88.85 & 85.48 & 87.26 & 84.77 & 88 & 89.87 & 86.76 \\ 
fog-3 & 90.69 & 78.07 & 85.2 & 80.07 & 83.32 & 78.94 & 83.77 & 86.82 & 82.78 \\ 
fog-4 & 86.72 & 69.56 & 78.92 & 72.27 & 75.89 & 71.01 & 77.96 & 81.56 & 75.62 \\ 
fog-5 & 68.6 & 49.04 & 53.9 & 52.33 & 52.88 & 49.68 & 57.67 & 62.29 & 55.18 \\ 
frost-1 & 89.97 & 83.66 & 84.7 & 84.07 & 85.76 & 83.64 & 85.51 & 87.75 & 84.85 \\ 
frost-2 & 84.42 & 77.88 & 78.97 & 77.47 & 78.96 & 77.26 & 79.14 & 81.4 & 79.16 \\ 
frost-3 & 74.85 & 67.67 & 69.3 & 67.14 & 68.76 & 66.03 & 69.58 & 72.54 & 70.14 \\ 
frost-4 & 73.32 & 65.93 & 67.14 & 65.97 & 68.52 & 65.37 & 67.93 & 71.44 & 69.41 \\ 
frost-5 & 62.13 & 55.02 & 56.67 & 55.62 & 56.77 & 54.26 & 57.69 & 61.11 & 59.88 \\ 
snow-1 & 89.26 & 84.58 & 85.44 & 84.59 & 86.43 & 85.07 & 85.95 & 87.82 & 85.67 \\ 
snow-2 & 78.96 & 72.01 & 73.37 & 73.14 & 73.42 & 72.43 & 73.65 & 78.05 & 73.84 \\ 
snow-3 & 82.85 & 75.6 & 76.84 & 75.74 & 76.95 & 75.74 & 77.76 & 79.98 & 75.95 \\ 
snow-4 & 80.29 & 70.84 & 72.56 & 71.93 & 72.39 & 71.25 & 72.97 & 76.12 & 72.16 \\ 
snow-5 & 74.94 & 63.85 & 66.94 & 65.9 & 65.71 & 64.24 & 66.29 & 70.33 & 66.53 \\ 
contrast-1 & 93.82 & 87.23 & 89.96 & 87.93 & 89.68 & 87.88 & 90.16 & 91.53 & 88.84 \\ 
contrast-2 & 90.53 & 76.02 & 84 & 77.27 & 82.1 & 76.3 & 82.8 & 86.02 & 81.37 \\ 
contrast-3 & 85.84 & 63.97 & 77.1 & 65.77 & 72.77 & 64.18 & 74.62 & 79.30 & 71.89 \\ 
contrast-4 & 75.08 & 44.07 & 62.37 & 47.35 & 55.57 & 44.94 & 59.87 & 65.72 & 55.14\\ 
contrast-5 & 29.36 & 20 & 25.67 & 20.18 & 22.18 & 21.04 & 25.28 & 25.66 & 24.04 \\ 
elastic\_transfor-1 & 89.97 & 82.97 & 84.5 & 83.38 & 84.74 & 82.48 & 84.42 & 86.54 & 83.25\\ 
elastic\_transfor-2 & 89.43 & 82.12 & 84.79 & 82.44 & 84.07 & 81.97 & 84.61 & 86.20 & 83.15 \\ 
elastic\_transfor-3 & 85.52 & 77.56 & 80.71 & 77.92 & 79.11 & 77 & 79.53 & 82.27 & 78.35 \\ 
elastic\_transfor-4 & 79.48 & 73.75 & 74.83 & 73.92 & 73.41 & 72.77 & 73.98 & 77.53 & 74.17 \\ 
elastic\_transfor-5 & 75.02 & 70.97 & 70.22 & 71.36 & 71.03 & 71.63 & 71.65 & 75.31 & 71.49 \\ 
jpeg\_copression-1 & 87.36 & 83.28 & 83.93 & 84.07 & 83.83 & 83.77 & 84.3 & 85.65 & 83.63 \\ 
jpeg\_copression-2 & 81.68 & 80.09 & 79.66 & 79.77 & 80.03 & 80.29 & 80.59 & 81.66 & 79.8 \\ 
jpeg\_copression-3 & 79.98 & 78.55 & 78.21 & 78.32 & 78.27 & 78.99 & 78.51 & 79.94 & 78.44 \\ 
jpeg\_copression-4 & 77.17 & 77.12 & 75.78 & 77.44 & 77.04 & 77.46 & 77.08 & 77.67 & 76.71 \\ 
jpeg\_copression-5 & 73.85 & 74.51 & 73.04 & 74.65 & 74.13 & 75.26 & 74.16 & 74.85 & 74.19 \\ 
pixelate-1 & 92.57 & 86.97 & 88.19 & 87.39 & 88.73 & 87.47 & 88.17 & 89.42 & 87.26 \\ 
pixelate-2 & 88.23 & 81.91 & 80.95 & 82.37 & 82.82 & 81.93 & 81.99 & 83.80 & 80.95  \\ 
pixelate-3 & 84 & 78.4 & 75.25 & 78.63 & 78.03 & 77.15 & 77.28 & 78.89  & 75.30 \\ 
pixelate-4 & 68.51 & 64.11 & 58.11 & 62.49 & 60.89 & 61.43 & 60.06 & 61.71 & 59.95 \\ 
pixelate-5 & 50.57 & 50.68 & 44.77 & 48.18 & 45.44 & 46.74 & 43.27 & 47.29 & 45.62 \\ 
saturate-1 & 92.41 & 84.98 & 88.38 & 85.38 & 87.26 & 85.39 & 88.23 & 89.34 & 86.82 \\ 
saturate-2 & 90.12 & 80.74 & 85.26 & 81.57 & 82.68 & 80.74 & 84.22 & 86.06 & 83.09 \\ 
saturate-3 & 93.83 & 87.89 & 90.45 & 88.12 & 89.53 & 88.03 & 90.15 & 90.97 & 88.73 \\ 
saturate-4 & 91.61 & 82.5 & 86.88 & 82.6 & 84.66 & 82.44 & 86.04 & 87.56 & 83.70 \\ 
saturate-5 & 87.48 & 76.03 & 82.76 & 75.53 & 78.3 & 75.85 & 80.62 & 82.64 & 77.00 \\ 
spatter-1 & 91.17 & 87.5 & 89.75 & 87.83 & 89.34 & 87.9 & 89.42 & 91.00 & 88.98  \\ 
spatter-2 & 85.2 & 83.85 & 85.98 & 83.52 & 85.64 & 84.72 & 85.88 & 87.59 & 85.00 \\ 
spatter-3 & 80.63 & 77.94 & 80.33 & 77.95 & 80.19 & 79.41 & 80.07 & 82.65 & 79.50 \\ 
spatter-4 & 94.68 & 84.57 & 86.71 & 84.77 & 86.51 & 85.14 & 86.72 & 88.22 & 85.32 \\ 
spatter-5 & 74.07 & 78.85 & 80.77 & 78.71 & 80.94 & 79.71 & 80.51 & 83.14 & 79.48 \\ 
\midrule
Overall & 74.11 & 69.43 & 71.51 & 69.36 & 70.76 & 69.09 & 70.31 & 73.56 & 70.70 \\ 
\bottomrule
\end{tabular}}
\end{table}

\begin{table}[h]
\renewcommand\arraystretch{1.2}
\centering
\caption{Sensitivity to Hyper Parameters in Training (1/2).}
\label{tab:appendix-sensitivity-acc1}
\vspace{-0.1in}
\setlength{\tabcolsep}{1.6mm}
{
\begin{tabular}{lccrrccccccc}
\toprule
{Algorithm} & {Epoch} & {Optimizer} & {Learning Rate} & {Scheduler} & {Acc.} & {Acc.$_{1}$} & {Acc.$_{2}$} & {Acc.$_{3}$} & {Acc.$_{4}$} & {mean} & std\\
\midrule
\multirow{8}{*}{FP32} & 200 & SGD & 0.1 & cosine & 94.58 & 94.6 & 95.05 & 94.64 & 94.84 & 94.74  & 0.20   \\
        ~ & 200 & SGD & 0.1 & step & 92.63 & 92.42 & 92.15 & 92.62 & 92.38 & 92.44  & 0.20   \\
        ~ & 200 & SGD & 0.01 & cosine & 92.23 & 91.76 & 91.76 & 91.99 & 92.17 & 91.98  & 0.22   \\
        ~ & 200 & SGD & 0.01 & step & 83.94 & 83.50 & 82.80 & 84.13 & 83.89 & 83.65  & 0.53   \\
        ~ & 200 & Adam & 0.001 & cosine & 93.51 & 92.94 & 93.12 & 93.35 & 92.86 & 93.16  & 0.27   \\
        ~ & 200 & Adam & 0.001 & step & 93.37 & 93.15 & 93.32 & 93.41 & 93.35 & 93.32  & 0.10  \\
        ~ & 200 & Adam & 0.0001 & cosine & 89.97 & 89.92 & 89.96 & 89.9 & 89.92 & 89.93  & 0.03   \\
        ~ & 200 & Adam & 0.0001 & step & 90.57 & 89.91 & 90.43 & 90.25 & 90.31 & 90.29  & 0.25   \\
\midrule
\multirow{8}{*}{BNN} & 200 & SGD & 0.1 & cosine & 87.62 & 87.53 & 87.99 & 88.86 & 87.84 & 87.97  & 0.53   \\
        ~ & 200 & SGD & 0.1 & step & 70.87 & 73.86 & 71.83 & 73.1 & 72.87 & 72.51  & 1.17   \\
        ~ & 200 & SGD & 0.01 & cosine & 73.52 & 72.62 & 72.82 & 71.14 & 72.59 & 72.54  & 0.87   \\
        ~ & 200 & SGD & 0.01 & step & 52.85 & 51.77 & 52.00 & 52.34 & 53.14 & 52.42  & 0.57   \\
        ~ & 200 & Adam & 0.001 & cosine & 88.76 & 88.99 & 88.67 & 88.84 & 88.81 & 88.81  & 0.12   \\
        ~ & 200 & Adam & 0.001 & step & 88.85 & 89.34 & 88.77 & 89.02 & 89.00 & 89.00  & 0.22   \\
        ~ & 200 & Adam & 0.0001 & cosine & 83.46 & 83.09 & 83.20 & 83.70 & 83.20 & 83.33  & 0.25   \\
        ~ & 200 & Adam & 0.0001 & step & 84.08 & 84.11 & 84.20 & 84.31 & 83.56 & 84.05  & 0.29   \\
\midrule
\multirow{8}{*}{XNOR} & 200 & SGD & 0.1 & cosine & 91.83 & 91.99 & 91.87 & 92.01 & 91.56 & 91.85  & 0.18   \\
        ~ & 200 & SGD & 0.1 & step & 90.02 & 90.01 & 90.12 & 89.82 & 89.7 & 89.93  & 0.17   \\
        ~ & 200 & SGD & 0.01 & cosine & 90.09 & 89.68 & 90.01 & 89.97 & 90.00 & 89.95  & 0.16   \\
        ~ & 200 & SGD & 0.01 & step & 86.86 & 86.66 & 87.21 & 86.98 & 86.61 & 86.86  & 0.24   \\
        ~ & 200 & Adam & 0.001 & cosine & 89.39 & 89.81 & 89.73 & 89.91 & 89.75 & 89.72  & 0.20   \\
        ~ & 200 & Adam & 0.001 & step & 89.92 & 89.79 & 89.73 & 90.01 & 89.61 & 89.81  & 0.16   \\
        ~ & 200 & Adam & 0.0001 & cosine & 86.18 & 86.29 & 87.03 & 86.36 & 86.62 & 86.50  & 0.34   \\
        ~ & 200 & Adam & 0.0001 & step & 86.32 & 87.04 & 86.68 & 86.99 & 87.18 & 86.84  & 0.34   \\
\midrule
\multirow{8}{*}{DoReFa} & 200 & SGD & 0.1 & cosine & 85.64 & 85.67 & 85.89 & 86.00 & 85.79 & 85.80  & 0.15   \\
        ~ & 200 & SGD & 0.1 & step & 86.95 & 86.98 & 86.69 & 86.62 & 86.65 & 86.78  & 0.17   \\
        ~ & 200 & SGD & 0.01 & cosine & 86.56 & 86.59 & 86.52 & 86.69 & 86.88 & 86.65  & 0.14   \\
        ~ & 200 & SGD & 0.01 & step & 78.76 & 79.97 & 80.73 & 79.94 & 80.47 & 79.97  & 0.76   \\
        ~ & 200 & Adam & 0.001 & cosine & 88.85 & 89.06 & 88.92 & 88.87 & 88.75 & 88.89  & 0.11   \\
        ~ & 200 & Adam & 0.001 & step & 89.08 & 89.16 & 88.93 & 89.23 & 88.84 & 89.05  & 0.16   \\
        ~ & 200 & Adam & 0.0001 & cosine & 83.56 & 83.17 & 83.65 & 83.60 & 83.66 & 83.53  & 0.20   \\
        ~ & 200 & Adam & 0.0001 & step & 83.70 & 83.74 & 84.27 & 84.19 & 84.01 & 83.98  & 0.26   \\
\midrule
\multirow{8}{*}{Bi-Real} & 200 & SGD & 0.1 & cosine & 87.55 & 87.81 & 88.06 & 87.30 & 87.88 & 87.72  & 0.30   \\
        ~ & 200 & SGD & 0.1 & step & 87.95 & 88.35 & 88.13 & 87.73 & 88.25 & 88.08  & 0.25   \\
        ~ & 200 & SGD & 0.01 & cosine & 87.76 & 87.93 & 87.73 & 87.72 & 87.64 & 87.76  & 0.11   \\
        ~ & 200 & SGD & 0.01 & step & 83.75 & 82.91 & 82.82 & 82.91 & 83.39 & 83.16  & 0.40   \\
        ~ & 200 & Adam & 0.001 & cosine & 88.78 & 89.15 & 89.06 & 89.00 & 89.2 & 89.04  & 0.16   \\
        ~ & 200 & Adam & 0.001 & step & 88.89 & 88.98 & 88.78 & 89.11 & 89.05 & 88.96  & 0.13   \\
        ~ & 200 & Adam & 0.0001 & cosine & 83.96 & 84.17 & 84.37 & 83.54 & 84.07 & 84.02  & 0.31   \\
        ~ & 200 & Adam & 0.0001 & step & 84.63 & 84.48 & 84.32 & 84.75 & 84.29 & 84.49  & 0.20   \\
\bottomrule
\end{tabular}}
\end{table}

\begin{table}[h]
\renewcommand\arraystretch{1.5}
\centering
\caption{Sensitivity to Hyper Parameters in Training (2/2).}
\label{tab:appendix-sensitivity-acc2}
\vspace{-0.1in}
\setlength{\tabcolsep}{1.6mm}
{
\begin{tabular}{lrrrrrrrrrrr}
\toprule
{Algorithm} & {Epoch} & {Optimizer} & {Learning Rate} & {Scheduler} & {Acc.} & {Acc.$_{1}$} & {Acc.$_{2}$} & {Acc.$_{3}$} & {Acc.$_{4}$} & {mean} & {std}\\
\midrule
\multirow{8}{*}{XNOR++} & 200 & SGD & 0.1 & cosine & 87.82 & 88.42 & 88.12 & 88.55 & 88.19 & 88.22  & 0.28   \\
        ~ & 200 & SGD & 0.1 & step & 73.55 & 73.11 & 75.06 & 74.05 & 73.78 & 73.91  & 0.73   \\
        ~ & 200 & SGD & 0.01 & cosine & 74.03 & 75.06 & 73.64 & 74.53 & 74.71 & 74.39  & 0.56   \\
        ~ & 200 & SGD & 0.01 & step & 53.55 & 54.16 & 54.01 & 52.91 & 54.36 & 53.80  & 0.58   \\
        ~ & 200 & Adam & 0.001 & cosine & 88.77 & 88.65 & 89.10 & 88.61 & 88.81 & 88.79  & 0.19   \\
        ~ & 200 & Adam & 0.001 & step & 89.18 & 89.05 & 89.27 & 88.93 & 89.00 & 89.09 & 0.14   \\
        ~ & 200 & Adam & 0.0001 & cosine & 83.86 & 83.49 & 83.56 & 83.16 & 83.62 & 83.54  & 0.25   \\
        ~ & 200 & Adam & 0.0001 & step & 83.46 & 83.77 & 84.40 & 84.06 & 83.82 & 83.90  & 0.35   \\
\midrule
\multirow{8}{*}{ReActNet} & 200 & SGD & 0.1 & cosine & 88.60 & 88.53 & 88.38 & 88.48 & 88.89 & 88.58  & 0.19   \\
        ~ & 200 & SGD & 0.1 & step & 88.42 & 88.01 & 88.10 & 88.02 & 88.43 & 88.20  & 0.21   \\
        ~ & 200 & SGD & 0.01 & cosine & 87.75 & 87.86 & 88.00 & 87.80 & 88.02 & 87.89  & 0.12   \\
        ~ & 200 & SGD & 0.01 & step & 83.29 & 82.89 & 83.65 & 83.76 & 83.27 & 83.37  & 0.35   \\
        ~ & 200 & Adam & 0.001 & cosine & 89.47 & 89.29 & 89.01 & 89.05 & 89.14 & 89.19  & 0.19   \\
        ~ & 200 & Adam & 0.001 & step & 89.27 & 89.74 & 89.48 & 89.40 & 89.39 & 89.46  & 0.18   \\
        ~ & 200 & Adam & 0.0001 & cosine & 84.65 & 84.93 & 84.48 & 84.65 & 84.67 & 84.68  & 0.16   \\
        ~ & 200 & Adam & 0.0001 & step & 84.69 & 84.55 & 84.93 & 84.94 & 85.38 & 84.90  & 0.32   \\
\midrule
\multirow{8}{*}{ReCU} & 200 & SGD & 0.1 & cosine & 91.72 & 91.94 & 91.68 & 91.69 & 91.81 & 91.77  & 0.11   \\
        ~ & 200 & SGD & 0.1 & step & 87.73 & 88.14 & 87.81 & 88.02 & 87.91 & 87.92  & 0.16   \\
        ~ & 200 & SGD & 0.01 & cosine & 87.32 & 87.28 & 87.53 & 87.48 & 87.32 & 87.39  & 0.11  \\
        ~ & 200 & SGD & 0.01 & step & 71.86 & 71.72 & 71.78 & 72.26 & 71.59 & 71.84  & 0.25   \\
        ~ & 200 & Adam & 0.001 & cosine & 88.24 & 89.98 & 88.26 & 88.48 & 88.13 & 88.62 & 0.77   \\
        ~ & 200 & Adam & 0.001 & step & 88.36 & 88.48 & 88.55 & 88.42 & 88.63 & 88.49  & 0.11  \\
        ~ & 200 & Adam & 0.0001 & cosine & 80.07 & 81.10 & 80.62 & 81.09 & 79.95 & 80.57  & 0.55   \\
        ~ & 200 & Adam & 0.0001 & step & 81.26 & 81.42 & 81.08 & 81.58 & 81.69 & 81.41  & 0.24   \\
\midrule
\multirow{8}{*}{FDA} & 200 & SGD & 0.1 & cosine & 89.69 & 89.59 & 89.56 & 89.53 & 89.65 & 89.60  & 0.07   \\
        ~ & 200 & SGD & 0.1 & step & 80.38 & 80.34 & 80.83 & 80.52 & 80.52 & 80.52  & 0.19   \\
        ~ & 200 & SGD & 0.01 & cosine & 80.72 & 80.93 & 80.89 & 80.70 & 80.79 & 80.81  & 0.10   \\
        ~ & 200 & SGD & 0.01 & step & 63.41 & 62.85 & 63.04 & 63.04 & 63.14 & 63.10  & 0.20   \\
        ~ & 200 & Adam & 0.001 & cosine & 89.70 & 89.57 & 89.57 & 89.80 & 89.76 & 89.68  & 0.11   \\
        ~ & 200 & Adam & 0.001 & step & 89.84 & 89.85 & 90.10 & 89.79 & 90.01 & 89.92  & 0.13   \\
        ~ & 200 & Adam & 0.0001 & cosine & 89.59 & 89.10 & 89.34 & 89.31 & 89.51 & 89.37  & 0.19   \\
        ~ & 200 & Adam & 0.0001 & step & 89.52 & 89.59 & 89.52 & 89.64 & 89.58 & 89.57  & 0.05   \\
\bottomrule
\end{tabular}}
\end{table}

\begin{table}[h]
\renewcommand\arraystretch{0.92}
\centering
\caption{Inference Efficiency on Hardware (1/4).}
\label{app:tab:eff1}
\vspace{-0.1in}
\setlength{\tabcolsep}{3.3mm}
{
\begin{tabular}{lllcrrrrrrrr}
\toprule
\multirow{2}{*}{Hardware} & \multirow{2}{*}{Threads} & \multirow{2}{*}{Arch.} & \multicolumn{3}{c}{Larq} & \multicolumn{2}{c}{daBNN} \\ 

& & & {FP32} & {BNN} & {ReAct} & {FP32} & {BNN}\\
\midrule

\multirow{13}{*}{Kirin 970} 
&\multirow{3}{*}{1} 
&   ResNet18  &  716.427 & 123.263 & 126.457 & 427.585 & 72.585  & \\
& & ResNet34  &  1449.67 & 159.615 & 171.227 & 836.321 & 124.091 & \\
& & VGG-Small &  242.443 & 14.833  & 16.401  & --      & --      & \\
\cmidrule{2-8}

&\multirow{3}{*}{2} 
&   ResNet18  &  372.642 & 72.697  & 78.605  & --      & --      & \\
& & ResNet34  &  732.355 & 96.711  & 108.41  & --      & --      & \\
& & VGG-Small &  121.91  & 10.304  & 11.935  & --      & --      & \\
\cmidrule{2-8}

&\multirow{3}{*}{4} 
&   ResNet18  &  191.517 & 42.986  & 47.182  & --      & --      & \\
& & ResNet34  &  367.891 & 61.413  & 73.101  & --      & --      & \\
& & VGG-Small &  57.721  & 8.72    & 8.387   & --      & --      & \\
\cmidrule{2-8}

& \multirow{3}{*}{8} 
&   ResNet18  &  96.937  & 37.457  & 56.017  & --      & --      & \\
& & ResNet34  &  212.982 & 53.809  & 67.667  & --      & --      & \\
& & VGG-Small &  33.647  & 18.649  & 19.818  & --      & --      & \\
\midrule

\multirow{13}{*}{Kirin 980}
&\multirow{3}{*}{1}
&   ResNet18  &  307.624 &   49.009 &   50.018 &  158.363 &   31.803 & \\
& & ResNet34  &  507.734 &   71.909 &   74.920 &  308.537 &   53.031 & \\
& & VGG-Small &   83.163 &    7.772 &    8.215 &    --   &    --   & \\
\cmidrule{2-8}

&\multirow{3}{*}{2}
&   ResNet18  &  187.494 &   52.057 &   54.285 &    --   &    --   & \\
& & ResNet34  &  367.853 &   57.336 &   60.483 &    --   &    --   & \\
& & VGG-Small &   49.264 &    6.116 &    5.604 &    --   &    --   & \\
\cmidrule{2-8}

&\multirow{3}{*}{4}
&   ResNet18  &  104.076 &   29.556 &   35.539 &    --   &    --   & \\
& & ResNet34  &  202.173 &   31.324 &   35.911 &    --   &    --   & \\
& & VGG-Small &   22.690 &    3.147 &    3.291 &    --   &    --   & \\
\cmidrule{2-8}

&\multirow{3}{*}{8}
&   ResNet18  &   60.307 &   45.683 &   56.416 &    --   &    --   & \\
& & ResNet34  &  120.738 &   60.758 &   86.887 &    --   &    --   & \\
& & VGG-Small &   18.147 &   21.688 &   23.350 &    --   &    --   & \\
\midrule

\multirow{13}{*}{Kirin 985}
&\multirow{3}{*}{1}
&   ResNet18  &  173.238 &   27.429 &   30.626 &  164.556 &   34.528 & \\
& & ResNet34  &  438.971 &   58.165 &   60.885 &  323.439 &   57.808 & \\
& & VGG-Small &   70.797 &    6.147 &    6.796 &    --   &    --   & \\
\cmidrule{2-8}

&\multirow{3}{*}{2}
&   ResNet18  &  103.621 &   25.672 &   35.477 &    --   &    --   & \\
& & ResNet34  &  327.416 &   53.949 &   62.865 &    --   &    --   & \\
& & VGG-Small &   55.328 &    5.955 &    6.243 &    --   &    --   & \\
\cmidrule{2-8}

&\multirow{3}{*}{4}
&   ResNet18  &   92.387 &   26.728 &   34.778 &    --   &    --   & \\
& & ResNet34  &  184.050 &   39.881 &   52.153 &    --   &    --   & \\
& & VGG-Small &   28.076 &    8.919 &   14.795 &    --   &    --   & \\
\cmidrule{2-8}

&\multirow{3}{*}{8}
&   ResNet18  &  130.972 &   82.772 &   89.766 &    --   &    --   & \\
& & ResNet34  &  227.504 &  119.586 &  143.958 &    --   &    --   & \\
& & VGG-Small &   44.339 &   34.034 &   43.816 &    --   &    --   & \\
\midrule

\multirow{13}{*}{Kirin 990}
&\multirow{3}{*}{1}
&   ResNet18  &  114.238 &   21.235 &   22.066 &  144.205 &   29.239 & \\
& & ResNet34  &  227.043 &   31.545 &   32.821 &  275.502 &   49.476 & \\
& & VGG-Small &   38.118 &    3.338 &    3.482 &    --   &    --   & \\
\cmidrule{2-8}

&\multirow{3}{*}{2}
&   ResNet18  &   59.329 &   13.911 &   14.179 &    --   &    --   & \\
& & ResNet34  &  116.822 &   23.452 &   22.770 &    --   &    --   & \\
& & VGG-Small &   20.055 &    2.080 &    2.194 &    --   &    --   & \\
\cmidrule{2-8}

&\multirow{3}{*}{4}
&   ResNet18  &   38.403 &   10.280 &   12.208 &    --   &    --   & \\
& & ResNet34  &   81.273 &   15.570 &   17.727 &    --   &    --   & \\
& & VGG-Small &   13.508 &    1.542 &    1.760 &    --   &    --   & \\
\cmidrule{2-8}

&\multirow{3}{*}{8}
&   ResNet18  &   37.703 &   25.360 &   31.365 &    --   &    --   & \\
& & ResNet34  &   78.753 &   34.884 &   39.363 &    --   &    --   & \\
& & VGG-Small &   12.707 &   14.414 &   21.749 &    --   &    --   & \\
\bottomrule

\end{tabular}}
\end{table}

\begin{table}[h]
\renewcommand\arraystretch{0.92}
\centering
\caption{Inference Efficiency on Hardware (2/4).}
\label{app:tab:eff2}
\vspace{-0.1in}
\setlength{\tabcolsep}{3.mm}
{
\begin{tabular}{lllcrrrrrrrr}
\toprule
\multirow{2}{*}{Hardware} & \multirow{2}{*}{Threads} & \multirow{2}{*}{Arch.} & \multicolumn{3}{c}{Larq} & \multicolumn{2}{c}{daBNN} \\ 

& & & {FP32} & {BNN} & {ReAct} & {FP32} & {BNN}\\
\midrule

\multirow{13}{*}{Kirin 9000E}
&\multirow{3}{*}{1}
&   ResNet18  &  118.059 &   19.865 &   20.547 &  129.270 &   24.781 & \\
& & ResNet34  &  236.047 &   31.822 &   32.575 &  250.680 &   42.134 & \\
& & VGG-Small &   39.114 &    3.595 &    3.832 &    --   &    --   & \\
\cmidrule{2-8}

&\multirow{3}{*}{2}
&   ResNet18  &   68.351 &   16.821 &   16.115 &    --   &    --   & \\
& & ResNet34  &  133.671 &   25.061 &   24.660 &    --   &    --   & \\
& & VGG-Small &   23.018 &    2.684 &    2.598 &    --   &    --   & \\
\cmidrule{2-8}

&\multirow{3}{*}{4}
&   ResNet18  &   45.592 &   17.452 &   18.847 &    --   &    --   & \\
& & ResNet34  &   91.648 &   23.395 &   28.022 &    --   &    --   & \\
& & VGG-Small &   14.360 &    2.762 &    2.782 &    --   &    --   & \\
\cmidrule{2-8}

&\multirow{3}{*}{8}
&   ResNet18  &   43.363 &   61.263 &   42.328 &    --   &    --   & \\
& & ResNet34  &   89.405 &   70.232 &   93.558 &    --   &    --   & \\
& & VGG-Small &   19.070 &   17.351 &   23.825 &    --   &    --   & \\
\midrule

\multirow{13}{*}{Dimensity 820}
&\multirow{3}{*}{1}
&   ResNet18  &  158.835 &   32.636 &   34.912 &  323.035 &   63.471 & \\
& & ResNet34  &  328.020 &   57.133 &   60.807 &  629.493 &  102.443 & \\
& & VGG-Small &   82.417 &    5.958 &    6.420 &    --   &    --   & \\
\cmidrule{2-8}

&\multirow{3}{*}{2}
&   ResNet18  &  122.167 &   29.306 &   34.384 &    --   &    --   & \\
& & ResNet34  &  250.088 &   43.306 &   50.143 &    --   &    --   & \\
& & VGG-Small &   51.320 &    4.670 &    5.053 &    --   &    --   & \\
\cmidrule{2-8}

&\multirow{3}{*}{4}
&   ResNet18  &   94.636 &   21.850 &   30.027 &    --   &    --   & \\
& & ResNet34  &  177.757 &   33.809 &   40.816 &    --   &    --   & \\
& & VGG-Small &   45.056 &    4.223 &    4.546 &    --   &    --   & \\
\cmidrule{2-8}

&\multirow{3}{*}{8}
&   ResNet18  &   90.210 &   45.357 &   61.981 &    --   &    --   & \\
& & ResNet34  &  166.989 &   68.444 &   74.286 &    --   &    --   & \\
& & VGG-Small &   32.971 &   21.344 &   23.706 &    --   &    --   & \\
\midrule

\multirow{13}{*}{Dimensity 9000}
&\multirow{3}{*}{1}
&   ResNet18  &  106.388 &   21.023 &   24.770 &  148.690 &   29.030 & \\
& & ResNet34  &  210.665 &   32.841 &   34.590 &  284.438 &   49.854 & \\
& & VGG-Small &   42.057 &    4.410 &    5.530 &    --   &    --   & \\
\cmidrule{2-8}

&\multirow{3}{*}{2}
&   ResNet18  &   81.606 &   22.661 &   27.050 &    --   &    --   & \\
& & ResNet34  &  143.349 &   27.666 &   37.910 &    --   &    --   & \\
& & VGG-Small &   26.512 &    2.273 &    2.410 &    --   &    --   & \\
\cmidrule{2-8}

&\multirow{3}{*}{4}
&   ResNet18  &   51.421 &   13.079 &   15.200 &    --   &    --   & \\
& & ResNet34  &  100.249 &   23.314 &   25.920 &    --   &    --   & \\
& & VGG-Small &   17.735 &    3.015 &    3.770 &    --   &    --   & \\
\cmidrule{2-8}

&\multirow{3}{*}{8}
&   ResNet18  &   43.355 &   24.939 &   30.740 &    --   &    --   & \\
& & ResNet34  &   84.182 &   30.212 &   39.990 &    --   &    --   & \\
& & VGG-Small &   14.857 &   14.258 &   17.540 &    --   &    --   & \\
\midrule

\multirow{13}{*}{Snapdragon 855+}
&\multirow{3}{*}{1}
&   ResNet18  &   90.430 &   19.769 &   20.530 &  163.293 &   31.174 & \\
& & ResNet34  &  186.694 &   29.126 &   30.512 &  298.882 &   49.948 & \\
& & VGG-Small &   29.735 &    3.153 &    3.259 &    --   &    --   & \\
\cmidrule{2-8}

&\multirow{3}{*}{2}
&   ResNet18  &   58.510 &   25.780 &   26.331 &    --   &    --   & \\
& & ResNet34  &  124.580 &   31.023 &   32.646 &    --   &    --   & \\
& & VGG-Small &   19.408 &    2.258 &    2.471 &    --   &    --   & \\
\cmidrule{2-8}

&\multirow{3}{*}{4}
&   ResNet18  &   39.269 &   19.865 &   23.297 &    --   &    --   & \\
& & ResNet34  &   82.180 &   30.248 &   31.387 &    --   &    --   & \\
& & VGG-Small &   13.566 &    2.032 &    2.359 &    --   &    --   & \\
\cmidrule{2-8}

&\multirow{3}{*}{8}
&   ResNet18  &   36.630 &   49.060 &   85.861 &    --   &    --   & \\
& & ResNet34  &   73.513 &   41.131 &   88.101 &    --   &    --   & \\
& & VGG-Small &   12.860 &   17.828 &   23.489 &    --   &    --   & \\
\bottomrule

\end{tabular}}
\end{table}

\begin{table}[h]
\renewcommand\arraystretch{0.92}
\centering
\caption{Inference Efficiency on Hardware (3/4).}
\label{app:tab:eff3}
\vspace{-0.1in}
\setlength{\tabcolsep}{2.7mm}
{
\begin{tabular}{lllcrrrrrrrr}
\toprule
\multirow{2}{*}{Hardware} & \multirow{2}{*}{Threads} & \multirow{2}{*}{Arch.} & \multicolumn{3}{c}{Larq} & \multicolumn{2}{c}{daBNN} \\ 

& & & {FP32} & {BNN} & {ReAct} & {FP32} & {BNN}\\
\midrule

\multirow{13}{*}{Snapdragon 870}
&\multirow{3}{*}{1}
&   ResNet18  &   88.145 &   16.527 &   17.020 &  126.762 &   25.240 & \\
& & ResNet34  &  185.468 &   25.488 &   26.195 &  237.361 &   41.440 & \\
& & VGG-Small &   30.318 &    2.851 &    2.964 &    --   &    --   & \\
\cmidrule{2-8}

&\multirow{3}{*}{2}
&   ResNet18  &   63.829 &   18.351 &   19.575 &    --   &    --   & \\
& & ResNet34  &  159.174 &   25.352 &   26.340 &    --   &    --   & \\
& & VGG-Small &   27.669 &    2.094 &    2.308 &    --   &    --   & \\
\cmidrule{2-8}

&\multirow{3}{*}{4}
&   ResNet18  &   42.796 &   17.578 &   21.083 &    --   &    --   & \\
& & ResNet34  &   89.960 &   25.816 &   27.201 &    --   &    --   & \\
& & VGG-Small &   14.829 &    2.614 &    2.215 &    --   &    --   & \\
\cmidrule{2-8}

&\multirow{3}{*}{8}
&   ResNet18  &   46.798 &   19.192 &   28.579 &    --   &    --   & \\
& & ResNet34  &   97.834 &   25.060 &   32.863 &    --   &    --   & \\
& & VGG-Small &   16.799 &    9.717 &   17.293 &    --   &    --   & \\
\midrule

\multirow{13}{*}{Snapdragon 888}
&\multirow{3}{*}{1}
&   ResNet18  &   77.205 &   15.547 &   16.111 &  123.618 &   25.240 & \\
& & ResNet34  &  152.887 &   22.906 &   23.893 &  234.972 &   41.648 & \\
& & VGG-Small &   25.133 &    2.410 &    2.543 &    --   &    --   & \\
\cmidrule{2-8}

&\multirow{3}{*}{2}
&   ResNet18  &   46.297 &   19.309 &   19.321 &    --   &    --   & \\
& & ResNet34  &   93.615 &   20.473 &   22.489 &    --   &    --   & \\
& & VGG-Small &   16.001 &    1.920 &    2.213 &    --   &    --   & \\
\cmidrule{2-8}

&\multirow{3}{*}{4}
&   ResNet18  &   33.524 &   13.699 &   14.332 &    --   &    --   & \\
& & ResNet34  &   67.495 &   19.020 &   21.157 &    --   &    --   & \\
& & VGG-Small &   11.743 &    2.882 &    2.768 &    --   &    --   & \\
\cmidrule{2-8}

&\multirow{3}{*}{8}
&   ResNet18  &   33.761 &   26.108 &   58.989 &    --   &    --   & \\
& & ResNet34  &   67.876 &   37.018 &   61.315 &    --   &    --   & \\
& & VGG-Small &   11.752 &   27.615 &   16.774 &    --   &    --   & \\
\midrule

\multirow{13}{*}{Raspberrypi 3B+}
&\multirow{3}{*}{1}
&   ResNet18  &  740.509 &  158.732 &  175.256 & 1460.723 &  241.713 & \\
& & ResNet34  & 1536.915 &  240.606 &  254.810 & 2774.888 &  435.170 & \\
& & VGG-Small &  257.079 &   24.479 &   25.790 &    --   &    --   & \\
\cmidrule{2-8}

&\multirow{3}{*}{2}
&   ResNet18  &  667.012 &  143.716 &  106.894 &    --   &    --   & \\
& & ResNet34  &  933.149 &  144.287 &  158.868 &    --   &    --   & \\
& & VGG-Small &  145.427 &   14.503 &   15.628 &    --   &    --   & \\
\cmidrule{2-8}

&\multirow{3}{*}{4}
&   ResNet18  &  562.567 &  108.585 &  116.640 &    --   &    --   & \\
& & ResNet34  &  976.223 &  159.258 &  183.698 &    --   &    --   & \\
& & VGG-Small &  191.470 &   10.839 &   10.196 &    --   &    --   & \\
\cmidrule{2-8}

&\multirow{3}{*}{8}
&   ResNet18  &  877.026 &  279.660 &  356.239 &    --   &    --   & \\
& & ResNet34  & 1638.035 &  389.924 &  485.260 &    --   &    --   & \\
& & VGG-Small &  399.338 &  110.448 &  142.978 &    --   &    --   & \\
\midrule

\multirow{13}{*}{Raspberrypi 4B}
&\multirow{3}{*}{1}
&   ResNet18  &  448.744 &   80.822 &   82.380 &  688.838 &  120.348 & \\
& & ResNet34  &  897.735 &  112.837 &  119.536 & 1362.893 &  209.276 & \\
& & VGG-Small &  150.814 &   11.177 &   12.024 &    --   &    --   & \\
\cmidrule{2-8}

&\multirow{3}{*}{2}
&   ResNet18  &  261.861 &   49.079 &   55.279 &    --   &    --   & \\
& & ResNet34  &  525.735 &   67.480 &   79.468 &    --   &    --   & \\
& & VGG-Small &   89.284 &    6.647 &    7.882 &    --   &    --   & \\
\cmidrule{2-8}

&\multirow{3}{*}{4}
&   ResNet18  &  270.191 &   36.331 &   45.903 &    --   &    --   & \\
& & ResNet34  &  572.423 &   53.866 &   70.841 &    --   &    --   & \\
& & VGG-Small &   90.650 &    5.056 &    6.167 &    --   &    --   & \\
\cmidrule{2-8}

&\multirow{3}{*}{8}
&   ResNet18  &  466.585 &  168.844 &  226.771 &    --   &    --   & \\
& & ResNet34  &  879.375 &  264.638 &  319.789 &    --   &    --   & \\
& & VGG-Small &  216.439 &  114.064 &  162.118 &    --   &    --   & \\
\bottomrule

\end{tabular}}
\end{table}

\begin{table}[h]
\renewcommand\arraystretch{1.5}
\centering
\caption{Inference Efficiency on Hardware (4/4).}
\label{app:tab:eff4}
\vspace{-0.1in}
\setlength{\tabcolsep}{3.4mm}
{
\begin{tabular}{lllcrrrrrrrr}
\toprule
\multirow{2}{*}{Hardware} & \multirow{2}{*}{Threads} & \multirow{2}{*}{Arch.} & \multicolumn{3}{c}{Larq} & \multicolumn{2}{c}{daBNN} \\ 

& & & {FP32} & {BNN} & {ReAct} & {FP32} & {BNN}\\
\midrule

\multirow{16}{*}{Apple M1}
&\multirow{3}{*}{1}
&   ResNet18  &   44.334 &    8.219 &    8.355 &    --   &    --   & \\
& & ResNet34  &   88.334 &   12.505 &   12.771 &    --   &    --   & \\
& & VGG-Small &   14.093 &    1.446 &    1.465 &    --   &    --   & \\
\cmidrule{2-8}

&\multirow{3}{*}{2}
&   ResNet18  &   24.775 &    5.037 &    5.194 &    --   &    --   & \\
& & ResNet34  &   47.179 &    7.425 &    7.690 &    --   &    --   & \\
& & VGG-Small &    7.398 &    0.829 &    0.854 &    --   &    --   & \\
\cmidrule{2-8}

&\multirow{3}{*}{4}
&   ResNet18  &   18.612 &    3.448 &    3.671 &    --   &    --   & \\
& & ResNet34  &   27.515 &    4.965 &    5.254 &    --   &    --   & \\
& & VGG-Small &    4.294 &    0.526 &    0.551 &    --   &    --   & \\
\cmidrule{2-8}

&\multirow{3}{*}{8}
&   ResNet18  &   16.653 &    5.035 &    6.003 &    --   &    --   & \\
& & ResNet34  &   27.680 &    6.445 &    6.953 &    --   &    --   & \\
& & VGG-Small &    3.996 &    0.735 &    0.712 &    --   &    --   & \\
\cmidrule{2-8}

&\multirow{3}{*}{16}
&   ResNet18  &   90.323 &   70.697 &   73.729 &    --   &    --   & \\
& & ResNet34  &  162.057 &  130.907 &  125.362 &    --   &    --   & \\
& & VGG-Small &   25.366 &   23.050 &   23.194 &    --   &    --   & \\
\midrule

\multirow{16}{*}{Apple M1 Max}
&\multirow{3}{*}{1}
&   ResNet18  &   46.053 &    8.653 &    8.486 &    --   &    --   & \\
& & ResNet34  &   91.861 &   12.593 &   13.039 &    --   &    --   & \\
& & VGG-Small &   14.285 &    1.454 &    1.488 &    --   &    --   & \\
\cmidrule{2-8}

&\multirow{3}{*}{2}
&   ResNet18  &   25.039 &    5.450 &    5.361 &    --   &    --   & \\
& & ResNet34  &   51.860 &    7.579 &    8.925 &    --   &    --   & \\
& & VGG-Small &    7.657 &    0.855 &    0.896 &    --   &    --   & \\
\cmidrule{2-8}

&\multirow{3}{*}{4}
&   ResNet18  &   14.708 &    3.625 &    3.888 &    --   &    --   & \\
& & ResNet34  &   27.933 &    5.266 &    6.021 &    --   &    --   & \\
& & VGG-Small &    4.292 &    0.576 &    0.620 &    --   &    --   & \\
\cmidrule{2-8}

&\multirow{3}{*}{8}
&   ResNet18  &   10.660 &    3.718 &    4.510 &    --   &    --   & \\
& & ResNet34  &   18.988 &    4.745 &    5.457 &    --   &    --   & \\
& & VGG-Small &    3.432 &    0.560 &    0.629 &    --   &    --   & \\
\cmidrule{2-8}

&\multirow{3}{*}{16}
&   ResNet18  &   60.500 &   47.727 &   53.900 &    --   &    --   & \\
& & ResNet34  &  120.449 &   91.464 &   96.356 &    --   &    --   & \\
& & VGG-Small &   21.354 &   13.868 &   15.311 &    --   &    --   & \\
\midrule

\end{tabular}}
\end{table}

\begin{table}[!t]
\centering
\renewcommand\arraystretch{3}
\caption{Accuracy and efficiency comparison among multi-bit quantization (2\&8-bits), pruning, and binarization.}
\vspace{-0.1in}
\label{app:tab:other_tech}
\resizebox{\linewidth}{!}
{\begin{tabular}{lllll}
\toprule
~ & \multicolumn{2}{c}{Accuracy} & \multicolumn{2}{c}{Efficiency} \\
~ & CIFAR10-Res18 & CIFAR10-Res20 & FLOPs (Res20, G) & Param (Res20, K) \\ \midrule
FP32 & 94.82 & 91.99 & 13.61 & 11690 \\ 
\midrule
Binary (overall) & 91.08 (-3.74) & 85.74 (-6.25) & 1.105 (12.32$\times$) & 884 (13.22$\times$) \\
\midrule
DoReFa-INT2~\citep{Dorefa-Net} & 92.71 & 87.56 & 1.686 & 1681 \\ 
PACT-INT2~\citep{PACT} & 92.98 & 88.12 & 1.686 & 1681 \\ 
LSQ-INT2~\citep{esser2019learned} & 93.11 & 89.26 & 1.686 & 1681 \\ 
INT2 (overall) & 92.93 (-1.89) & 88.31 (-3.68) & 1.686 (8.07$\times$) & 1681 (6.95$\times$) \\ 
\midrule
DoReFa-INT8~\citep{Dorefa-Net} & 94.79 & 91.83 & 7.278 & 4067 \\ 
PACT-INT8~\citep{PACT} & 94.8 & 91.87 & 7.278 & 4067 \\ 
LSQ-INT8~\citep{esser2019learned} & 94.78 & 91.95 & 7.278 & 4067 \\ 
INT8 (overall) & 94.79 (-0.03) & 91.88 (-0.11) & 7.278 (1.87$\times$) & 4067 (2.87$\times$) \\ 
\midrule
\citep{li2016pruning} & 94.47 & 91.32 & 9.527 & 9936 \\ 
ResRep~\citep{ding2021resrep} & 94.53 & 91.76 & 6.805 & 7247 \\ 
Pruning (overall) & 94.50 (-0.32) & 91.54 (-0.45) & 8.166 (1.67$\times$) & 8591 (1.36$\times$) \\ \bottomrule
\end{tabular}}
\end{table}

\subsection{Comparison Results against Other Compression Technologies}

We further evaluated representative multi-bit quantization algorithms~\citep{Dorefa-Net,PACT,esser2019learned} (with INT2 and INT8) and dropout (pruning) algorithms~\citep{li2016pruning,ding2021resrep} in the limited time to demonstrate their accuracy and efficiency metrics and compare them to network binarization. The results show that compared with multi-bit quantization and dropout, binarization brings more significant compression and acceleration while facing greater challenges from the decline in accuracy.

To highlight the characteristics of network binarization, we compare it with other mainstream compression approaches, including multi-bit quantization and pruning, from accuracy and efficiency perspectives (Table~\ref{app:tab:other_tech}). Overall, the results express the intuitive conclusion that there is a trade-off between accuracy and efficiency for different compression approaches. The ultra-low bit-width of network binarization brings acceleration and compression beyond multi-bit quantization and pruning. For example, binarization achieves 12.32$\times$ FLOPs saving, while INT8 quantization achieves 1.87$\times$. However, binarization algorithms also introduce a significant performance drop, the largest among all compression approaches (\textit{e.g.}, CIFAR10-Res20 binary 85.74 \textit{vs.} pruning 91.54). These results show that network binarization pursues a more radical efficiency improvement among existing compression approaches and is oriented for deployment on edge devices, which is consistent with the conclusions in BiBench.


\end{document}